\documentclass[11pt]{article}
\ifdefined\XeTeXversion\else
  \DeclareUnicodeCharacter{2192}{\ensuremath{\rightarrow}}
\fi
\usepackage[letterpaper,margin=1in]{geometry}
\usepackage{newtx}
\usepackage{microtype}
\usepackage{booktabs}
\usepackage{array}
\usepackage{graphicx}
\usepackage{longtable}
\usepackage{calc}
\usepackage{seqsplit}
\usepackage[round]{natbib}
\usepackage{caption}
\usepackage[colorlinks=true,linkcolor=black,citecolor=blue,urlcolor=blue]{hyperref}
\usepackage{xurl}
\usepackage{needspace}

\title{\bfseries One note in three: a verified census of three deployed AI scribes, and the instrument that counted it}
\author{Sebastian Fox\thanks{Lead and corresponding author: \texttt{seb@composo.ai}, ORCID \mbox{\href{https://orcid.org/0000-0001-9839-6952}{0000-0001-9839-6952}}.} \and Luke Markham \and Ryan Lail \and Michael Karotsieris}
\date{}

\newcommand{\partdiv}[1]{\needspace{7\baselineskip}\vspace{2.2\baselineskip}\par\noindent\rule{\linewidth}{0.4pt}\par\vspace{0.5\baselineskip}\noindent{\Large\bfseries #1}\par\nobreak\vspace{0.3\baselineskip}}
\providecommand{\tightlist}{\setlength{\itemsep}{0pt}\setlength{\parskip}{0pt}}

\begin{document}
\maketitle

\begin{abstract}
Ambient AI scribes draft clinical notes at scale, and the reassurance offered is that a clinician signs every note. We audited three commercial scribe products on the same 142 consultations - 565 notes across recorded UK primary-care and US ambulatory encounters plus authored scenarios. Twelve discovery passes proposed 13,678 candidate errors; the 5,898 that cleared an importance filter went to an adversarial panel of two AI models from different families, each instructed to refute every candidate it defensibly could, and 618 findings survived. One note in three (31.3\% {[}27.0, 35.6{]}) contains at least one verified failure, and the failures concentrate in allergy and medication information, invented patient identity, and - on telephone consultations that could contain none - history written up as physical examination. No product here was given a patient record, and with the two classes a record would have prefilled set aside, invented identities and invented dates, the rate is 24.8\% {[}20.8, 29.0{]}. The findings organise into a three-tier taxonomy whose classes come from the published scribe-error taxonomies, plus one failure mode it could not place: a treatment the clinician explicitly retracts recorded as delivered care, named from a single consultation across two products. Two clinicians adjudicated blind samples that did not overlap. A physician author upheld 20 of 21 verified findings (95.2\% {[}77.3, 99.2{]}), and an independent clinician, not an author and with no involvement in the study, upheld 12 of 12 ({[}75.8, 100{]}). Both judged every sampled refusal genuine. A failure rate, however, depends on the instrument that counted it as much as on the scribes, and we measured its share. With the model, the evidence and every setting held fixed, the review instruction alone moves the share of candidates verified from 9.3\% to 79.0\%, and the reviewing model family moves the headline: run alone at the same strict instruction, the gentler of the two families flags 54.8\% of the sampled notes against the harsher one's 27.8\%, roughly double. Adding it to the harsher family as a second opinion, with a tiebreak, adds one percentage point. Between 28\% and 97\% of sampled notes carry a verified failure depending on the standard applied. Published audits disagree with one another by a margin that instrument differences alone can produce: omission is 54--86\% of errors across them, ours is 23.1\%, and the one study that counts over notes as we do reports omission in 18\% of them where we find 15.4\%. The census shows a signing clinician where attention matters most, and gives a buyer the sharper question: an error rate, under what instrument? We release all 618 findings with transcript-side evidence, every prompt and model version, and the re-runnable pipeline, so the census can be repeated on other products under the same published standard.

\end{abstract}

\section{Introduction}

A telephone consultation contains no physical examination. Two of the three commercial ambient scribes we audited nonetheless produced notes for telephone consultations that document one: history rewritten as performed examination, in the part of the note a clinician signs and a later reader treats as observed fact. On one such consultation the clinician says out loud that they would normally examine at this point and cannot; the note that came back carries a physical-examination section describing the rash (Scribe C).

This paper is a census of such failures. Three commercial ambient scribes wrote notes from the same 142 consultations - recorded UK primary-care consultations, US ambulatory encounters, and two authored strata we wrote ourselves so that part of the corpus would have ground truth we controlled, 565 notes in all. Twelve separate discovery passes over every note proposed 13,678 candidate errors, and every candidate that cleared an importance filter was put before an adversarial verification panel: two models from different families, each shown the full note and full transcript and told to refute the finding if refuting it was at all defensible. 618 findings survived. The notes were produced by running these consultations through the deployed products, not captured from live clinical traffic - which is what makes it possible to put identical consultations through competing products, and means no real patient appears anywhere in the study.

The census is the first half of the paper. The second half is what the census teaches about failure rates themselves. A failure rate is a joint property of the scribes and of the instrument that counted it. Here that instrument is the twelve discovery passes and the verification panel, which together turn a set of notes into a count. We measured its contribution on the census's own material: a stratified fifth of the notes, drawn whole, was re-reviewed - 1,295 of the 5,898 candidates - with the model, the evidence and every setting unchanged, and the strict verification instruction replaced by a lenient one. The strict instruction verifies 9.3\% of those candidates, and the lenient one 79.0\%. The rest of the machinery - the second model family, the tiebreak - moves the count by one percentage point over the harsher reviewer acting alone. Within the review layer, almost the whole of the effect on the count is the standard the reviewer is told to apply rather than the panel's architecture; the discovery layer's own contribution is not measured here (Section 4.4). None of the human-review audits we compare against reports the effect of its own review standard on its counts - and Section 4 shows their widely divergent headline numbers are the size an instrument difference alone can produce, and that where one study counts the way we count, we match. So we publish both: the rates, and the complete instrument that produced them, re-runnable by anyone.

The paper makes four contributions. (i) A verified, evidence-quoted census and failure taxonomy of three deployed scribe products on a shared corpus, with cross-product replication measured and two clinicians' blinded review of the instrument's judgement layers. (ii) A controlled decomposition of the counting instrument's effect on the count - the review instruction, the model family and the panel architecture separately, and which claims survive a strict standard and why - with the instrument released and re-runnable. (iii) Evidence that the divergent error mixes of the published scribe audits are consistent with instrument differences of the size we measure. (iv) The practical reading, for the clinician who signs and the buyer who compares. A companion paper, released concurrently, takes up the question this census raises about automated quality layers: whether the LLM judges deployed over scribe output can detect the class of failure that published audits find dominant (``LLM Judges Verify Presence, Not Absence: Omission Blindness in AI Clinical Notes and What Recovers It'').

\section{Related work}

Five literatures meet here.

\textbf{Human-review audits of scribe output.} These establish the error classes and omission's dominance. Biro et al.~had residents read scripted encounters into two commercial products and reviewers categorise 127 errors under a four-class scheme developed as a patient-safety classification for generative AI \citep{hose2025development}: omission was 83\% of one product's errors and 54\% of the other's \citep{biro2025accuracy}. Anderson et al.~played 14 simulated encounters to five platforms and found omission 76.3\% {[}70.0, 83.3{]} of errors \citep{anderson2025evaluating}; Kernberg et al., prompting ChatGPT-4 over encounter transcripts rather than testing a deployed scribe, found 86.3\% \citep{kernberg2024using}; and Taylor et al.'s real-world pilot had physicians self-evaluate 356 of their own AI-drafted notes, with accidental omission the most prevalent concern, in 18\% of notes \citep{taylor2026quality}. Smaller audits repeat the shape: omission at 71\% of errors across four commercial tools \citep{arko2025documenting}, deletion and omission the common failures in a six-product primary-care comparison whose every product still rated good to excellent \citep{ha2025evaluating}, omissions predominating in a UK simulated audit \citep{draper2026ai}. All of these use human reviewers applying their own thresholds, and none of the four we compare against directly reports what its review standard contributed to its counts - so the omission shares above span 54 to 86\% of errors with no way to say how much of the spread is scribe and how much is standard. Our census differs in both respects: every finding is adversarially verified and carries its evidence, and the counting instrument is itself measured, decomposed and released.

\textbf{Shared-corpus comparisons across products.} The largest to date put eleven scribe products and eighteen human note-takers through five standardised primary-care cases for thirty blinded raters, and found AI notes below human notes on every domain of a modified PDQI-9, a note-quality rating scale \citep{reddy2026rapid}; a three-arm randomised deployment of two products measured physician experience across 48,349 scribe-arm visits, with errors reported by survey rather than note audit \citep{lukac2025ambient}. These designs compare quality scores or clinician experience; none produces a verified account of what fails. The instrument this line leans on hardest also carries measured reliability problems on exactly this material: on 220 emergency department scribe and non-scribe notes rated twice, PDQI-9 inter-rater agreement was a Pearson r of 0.07 and scores moved with the rater rather than the note \citep{walker2017pdqi}, and the one ambient-scribe evaluation framework to report a sensitivity analysis on its own scale found evaluator agreement changed when its five-point ratings became three-point \citep{wang2025evaluation}.

\textbf{Vendor self-evaluation.} Vendors have begun publishing evaluations of their own products. A vendor-authored, peer-reviewed comparison reports hallucinations in 31\% of its ambient notes against 20\% of physician-written notes for the same encounters, measured by a single binary rating item \citep{palm2025assessing} - and a standard that flags a fifth of physician-written notes is characterising itself as much as the product. A vendor preprint posted days before this paper was completed compares one product against clinician-written notes on 385 paired consultations across five countries, reports its error burden under two detectors of deliberately different sensitivity - 24.4\% of AI notes under a calibrated automated reviewer, 6.2\% under unaided clinician adjudication - and bounds rather than point-estimates the rate \citep{bergman2026quality}. Another vendor's evaluation attributes much of its measured incompleteness to overly strict alignment constraints in its own evaluation model \citep{hansen2025factsr}. And the largest sentence-level annotation of LLM-written notes - 50 doctors over 12,999 note sentences, finding hallucinations in 1.47\% of note sentences and omissions in 3.45\% of transcript sentences - is likewise vendor-affiliated work \citep{asgari2025framework}. This line is increasingly instrument-aware, and it is not independent: none of the vendor evaluations cited here publishes the instrument behind its numbers, and a favourable finding grades the grader's own product. This census evaluates no product of its authors, and releases its instrument in full: every prompt, model version and verdict.

\textbf{The review standard as the moving part.} That the standard moves the count has been arriving from several directions at once. The PriMock57 authors proposed consultation checklists in 2022 because expert evaluators of generated notes disagreed substantially \citep{savkov2022consultation}; a 2026 scoping review finds scribe validation methods highly heterogeneous, leaving cross-system comparison limited \citep{kerimoglu2026validating}; an editorial asks outright whether scribe measurement is measuring what matters \citep{coiera2026scribes}; and in the MHRA's regulatory sandbox, an LLM judge evaluating synthetic radiology impressions conflated omissions with a failure of groundedness where radiologists treated the two as distinct \citep{mhra2025airlock}. Two 2026 results convert the concern into counts, and both come from vendors' own research teams: one moves the flagged rate on the same structured progress notes (SOAP format) by a factor of nearly four by relabelling clinically reasonable inference as supported \citep{vachhani2026beyond}, and the preprint above bounds its error burden between two detectors of deliberately different sensitivity \citep{bergman2026quality}. The first argues that one of its two regimes is the correct one; the second bounds a rate between two, and Section 4.1 sets both beside our own. Our position is a third: neither standard is privileged, a rate is readable only next to the instrument that produced it, and so the instrument itself - what it keeps, what it refuses, and why - is a first-class object of this paper, published with the count.

\textbf{Fact enumeration, and the companion paper.} Fact-enumeration methods for measuring omissions in medical summaries begin with MED-OMIT, published at ML4H \citep{schumacher2025medomit}, which our verification and audit design is downstream of in spirit. Automated checking of what a note does contain is now close to clinician agreement: on a benchmark of the statements a generated hospital-course summary makes, the best configuration of an AI checker holding the patient's record agreed with clinicians on 93.2\% of them {[}92.3, 94.0{]} \citep{chung2026verifact}. The absence side - the reliability of LLM judges on what a note leaves out, the automated version of the sign-off this census argues is overloaded - is the companion paper's subject, with the wider judge literature reviewed there.

\textbf{Against that literature, our contribution} is the census published together with its instrument: to our knowledge the first adversarially verified, evidence-quoted account of what deployed scribes get wrong on identical consultations across three products, and a controlled decomposition of the counting instrument itself, weighing the review instruction, the model family and the panel architecture one against another. Both are released in full and independent of any vendor, so either can be re-run by anyone on any product. We start with what the products got wrong, because it is what makes the instrument question matter.

\section{What three deployed scribes get wrong}

\subsection{How the census was built: wide discovery, adversarial verification}

\textbf{The corpus.} The corpus pairs each of the 142 consultations with its transcript and the notes the three products wrote for it: 282 notes from Scribe A (generated through an API at two note templates per consultation), 141 from Scribe B and 142 from Scribe C. By source, 57 consultations are recorded UK primary care (PriMock57, real clinicians with actor patients, used in full) and 45 are US ambulatory encounters (ACI-Bench) \citep{papadopoulos-korfiatis-etal-2022-primock57,yim2023acibench}. The ACI-Bench encounters were a seeded draw of 48, stratified by the corpus's own split and by transcript length, from the 140 encounters outside its official training split, which we had used to develop an earlier tool and so excluded; three of the 48 were dropped in the ground-truth audit of their fact sheets and this stratum's notes were generated from the kept list, so 45 remain. PriMock57's own four audit drops fell the other way, because its notes were generated from all 57 and a fact sheet gates only the analyses that use one. We wrote the remaining 40 ourselves: 30 scenarios seeded with documentation traps - details of the kind scribes tend to mishandle, planted so that part of the corpus would have ground truth we controlled - and 10 that an author of this study wrote without knowing the trap scheme existed. The 565 notes divide 228, 180, 120 and 37 across those four sources. That last stratum is the check on whether our own trap-writing inflated the rates: the trap-seeded notes do run higher (0.475 findings per note against 0.297), but the difference is not distinguishable from zero at these sample sizes (Appendix A.6). The authored strata are not where the headline comes from: notes with at least one verified failure run 43.4\% on PriMock57 (99 of 228) and 28.3\% on ACI-Bench (51 of 180) against 18.3\% on the trap-seeded scenarios (22 of 120) and 13.5\% on the trap-blind ones (5 of 37), so removing everything we wrote would raise the pooled rate, not lower it. One consultation in the blind stratum produced a note from Scribe C only - the other two products' notes were never captured for it, which is missing data rather than a product failure and is counted as neither - so Scribes A and B cover 141 of the 142 consultations. Seven further notes on disk are excluded and named in the release: four capture probes from an audio speed-and-silence comparison, and three Scribe C notes for the three excluded ACI-Bench encounters, which that product's capture run had covered.

Each product received the consultation alone - replayed audio for Scribes B and C, the transcript through an API for Scribe A - with no patient record, demographics or encounter date beyond what the product's own application supplies. Integrated deployments often prefill some of that context, a difference Sections 3.3 and 3.4 return to. Most of the notes were captured between 9 and 11 August 2026, days before the discovery passes ran on the 12th. The remaining 121 - 84 of Scribe A's, 22 of Scribe B's and 15 of Scribe C's - come from a pilot capture in June 2026 and were carried into the corpus unchanged rather than regenerated, so the corpus holds two product snapshots six weeks apart rather than one. No version identifier was available to us from any of the three products, so the corpus is dated rather than versioned, and at two dates rather than one; Section 7 says what that means for replication.

\textbf{Discovery and verification.} Discovery is deliberately over-inclusive: eleven targeted passes, each hunting one known failure mode across every note, plus a twelfth open pass for anything we did not think to look for, all run on \texttt{anthropic/claude-opus-5}. Together they proposed 13,678 candidate errors, of which 5,898 cleared the discovering model's own importance filter. Verification is deliberately hostile: two skeptics from different model families - \texttt{anthropic/claude-opus-5}, which also ran discovery and is the harsher of the two, and \texttt{openai/gpt-5.5}, the gentler - each read the full note and the full transcript, and refute the candidate if refuting it is at all defensible. Later sections call them the harsher and the gentler skeptic, and their model families the harsher and gentler family. A finding counts if both keep it, or if they split and a third model, \texttt{openai/gpt-5.4} at high reasoning effort, upholds it; a failed or unparseable reply counts as a refutation. The harsher skeptic's calls ran at temperature 1.0 with medium reasoning effort, a 6,000-token cap and deterministic per-call seeds; exact pins and settings for every role are in the run manifests (Appendix A.3).

The panel verified 618 of the 5,898 - 10.48\% {[}8.9, 12.1{]}, or {[}9.72, 11.29{]} treating candidates as independent. Throughout this paper, figures in square brackets are 95\% confidence intervals. Findings are not merged across passes, so one underlying error surfaced by two hunts counts twice; finding counts are therefore counts of verified findings rather than of distinct errors, and the headline rate is stated over notes, where that distinction cannot inflate anything. A model pass grouped each note's findings by whether one correction would resolve them together, and a sample of its groupings was hand-checked; on that grouping, the 618 verified findings are approximately 265 distinct errors; a model-free grouping by description similarity reads 352 to 529 over the same findings, so the deduplication factor is between about 1.2 and 2.3 depending on how a repeat is defined (Appendix A.9). Each finding carries its evidence quotes, the panel's verdicts, an importance rating from the panel (recorded as salience in the released data and the appendix) and, for all but one, a severity grade against a written clinical rubric.

\textbf{The human checks.} A physician author blind-adjudicated the instrument's output and both of its judgement layers in structured sittings. The precision sitting covered the verified findings themselves. It held 30 items: 21 findings drawn uniformly at random from the 618, interleaved with 9 candidates the panel had refused - drawn from its high-importance refusals, so they were not trivially distinguishable from survivors - shuffled and stripped of every status label, each presented with its full note and full transcript. 20 of the 21 verified findings stood, a precision of 95.2\% (Wilson {[}77.3, 99.2{]}) for the published set. They also judged all nine refused candidates genuine failures. That does not contradict their endorsement of the panel's refusals in the refusal-and-rubric sitting below. The panel refuses on the grounds its strict instruction sets (scope, defensible inference, documentation convention, immateriality) and, in a quarter of the audited sample, because the fact is stated elsewhere in the note (Section 4.3), not because nothing is wrong with the note. These nine were drawn from candidates the panel had itself rated important and refused anyway, and its recorded reasons for them cite documentation convention and immateriality alike, so what the sitting tests is the strict end of the panel's standard rather than its low-salience discards. A refused candidate can be a genuine failure the census declined to count under its standard, and Section 4.1 reads the nine that way.

The refusal-and-rubric sitting, also the physician author's, covered the panel's refusals and the severity rubric together. Ten sampled refusals went in with their audit labels concealed; the physician author judged eight and abstained on two, calling none a wrongly refused omission. Twenty items were graded blind against the same written rubric: facts drawn from the companion paper's omission benchmark, where each test note has a named fact deliberately missing at a known severity grade, so the twenty span all three grades by construction, and eight of them were drawn deliberately from facts on which the rubric's two arms had split and the tiebreak resolved downward. That gave 70\% exact agreement with every disagreement within one grade - and the salted eight are not what holds it there, since agreement ran higher on them than on the facts the two arms agreed on (Appendix A.4). It calibrates the grading axis rather than the individual grades on these 618 findings, since the graded material is benchmark facts and not census findings. In the precision sitting they also regraded the 20 findings they upheld (16 exact, 4 graded above the rubric, none below). Beyond the structured sittings they spot-checked verified findings across the wider set (Section 4, Appendix A.4).

\begin{figure*}[tp]
\centering
\includegraphics[width=\linewidth]{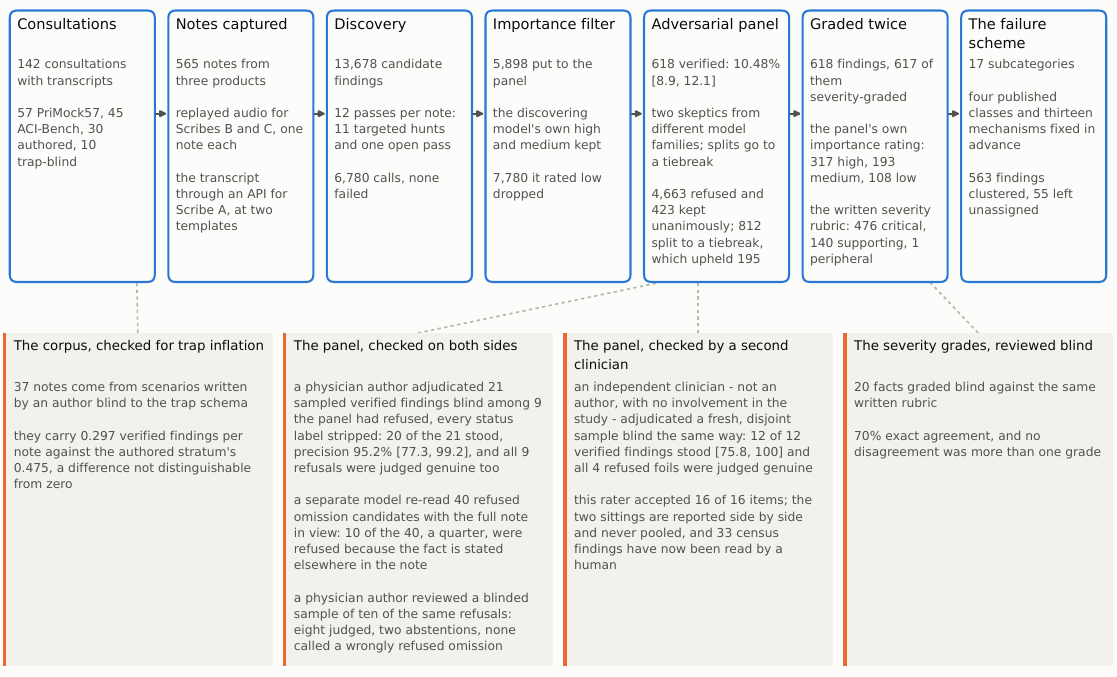}
\caption{\textbf{The census, end to end.} The upper band is what each stage produced; the lower band is the check that stands beside it and what that check found. 142 consultations from four sources were written up by three products under two capture paths, replayed audio for Scribes B and C and the transcript through an API at two note templates for Scribe A, giving 565 notes. Twelve discovery passes per note, eleven hunting a named failure mode and one open, proposed 13,678 candidates, of which 5,898 cleared the discovering model's own importance filter; two skeptics from different model families - both keeping, or a tiebreak settling a split - verified 618, 10.48\% [8.9, 12.1]. Survivors carry the panel's own importance rating and a severity grade against a written clinical rubric, which are two different instruments and are reported separately. Four cards sit in the lower band. Two of them put the panel's own output back to a clinician, the physician author's and the independent clinician's, on disjoint samples that are never pooled (Section~3.1); the physician author's card holds two further checks, a model auditor's re-read of 40 refused omission candidates and a blinded physician review of ten of those refusals. The severity totals count the clustered and unassigned findings together; Appendix A.1 splits them. Nothing in this figure ranks the products: the capture paths differ, and Scribe A contributes two notes per consultation where the others contribute one.}
\label{fig:pipeline}
\end{figure*}

A separate sitting put a fresh sample to an \textbf{independent clinician}: 16 items in two lots, 12 verified findings drawn uniformly at random from the 618 and 4 foils - candidates the panel had refused, drawn from its high-importance refusals - every one of them an item nobody had adjudicated before, with the precision sitting's items excluded and their absence asserted. The design and the wording were the precision sitting's. Independent here means not an author and with no involvement in the study, which is what the word means everywhere in this paper. All 12 verified findings stood, 12 of 12, Wilson {[}75.8, 100{]} over 11 consultations, and all four refused candidates were judged genuine too. \emph{This sitting and the precision sitting are reported side by side and never pooled}: the samples are disjoint and the raters are different people, so a combined figure would answer no question. What this sitting gains is coverage and a second pair of eyes - the number of census findings a human has assessed goes from 21 to 33 - and an interval running from 75.8\% to 100\% is wide by construction. Two limits go with it. This rater judged 16 of 16 items genuine and the physician author judged 29 of 30, so neither sitting can separate ``the flagged findings are nearly all real'' from ``clinicians asked to check flags accept them''; and because the samples are disjoint there is no interassessor agreement statistic to report, which is a deliberate design choice (Appendix A.4). The severity grades from this sitting run the other way from the precision sitting's, and Appendix A.1 reads the three gradings together. Figure~\ref{fig:pipeline} draws the pipeline end to end, with each stage's counts and the checks that sit beside it. The full protocol, prompts and parameters are in Appendix A.

\textbf{What the numbers estimate.} The primary estimand - the quantity the census's headline figures estimate - is the share of notes carrying at least one verified finding under the released discovery-and-verification instrument. Its intervals come from a bootstrap that resamples whole consultations (10,000 draws), because notes from one consultation share a transcript and are not independent. A candidate-level share is the fraction of proposed candidates that a given review standard verifies. It describes how the instrument behaves on a fixed pool of candidates, not how often the scribes err. Its headline intervals resample consultations in the same way, and where an interval instead treats notes or candidates as independent (a Wilson interval), the text says so. The contrasts between review standards in Section 4.1 resample consultations too, and paired contrasts between standards use an exact McNemar test, quoted as a bare p-value. Intervals and p-values throughout are unadjusted for multiplicity: the contrasts describe the instrument's behaviour rather than confirm hypotheses, and where a contrast below is reported with an interval that excludes zero, that is one unadjusted interval and no more. Neither quantity estimates the total number of errors in a note: an error no discovery pass proposes is invisible to both, so both can only undercount. Nor is the pooled share an estimate for any wider population of consultations: the four strata run from 13.5\% to 43.4\%, so the pooled figure carries the mixture we chose. The bootstrap draws whole consultations without regard to source, so that mixture varies from draw to draw rather than being held fixed; either way, the interval speaks only for a corpus assembled in these proportions.

\textbf{Quotation.} One convention governs every example in this paper. Transcript excerpts are quoted verbatim; the source corpora are CC BY 4.0. Note-side material appears either as a short verbatim fragment inside quotation marks or, at any greater length, as a paraphrase of our verified finding records, because the products' terms do not permit reproducing their notes at length. Attribution names the product letter.

\subsection{One note in three carries a verified failure}

Pooled across products, 177 of 565 notes carry at least one verified failure: 31.3\% {[}27.0, 35.6{]} (Table~\ref{tab:census}) - the interval resamples whole consultations, and treating notes as independent instead gives {[}27.6, 35.3{]}. That count is what survived two stages that can only push it down. Twelve discovery passes proposed the candidates; the panel then refuted roughly nine in ten. A real error the panel refuted and a real error no pass proposed are both missing from the total. So every note-level rate in this paper is a floor for the errors the instrument missed, though not for any it wrongly kept - the precision sitting of Section 3.1 bounds that direction, at 20 of 21 sampled findings upheld - and it is a floor under the standard this paper publishes. Section 4 measures which way that standard leans and by how much.

% T1 - the census. Generated by extract_p2_floats.py from master/findings_rates.json
% and master/findings_note_ci_clustered.json; do not edit by hand, regenerate.
% Requires \usepackage{booktabs}. Set as a full-width float: the interval column does not
% fit a two-column measure (in a single-column class table* behaves as table).
\begin{table*}[t]
\centering
\small
\setlength{\tabcolsep}{6pt}
\begin{tabular}{@{}lrrrll@{}}
\toprule
Product & notes & consultations & findings & findings per note (SD) & notes with at least one finding\\
\midrule
Scribe A & 282 & 141 & 128 & 0.454 (1.05) & 60 = 21.3\% [16.0, 27.0] \\
Scribe B & 141 & 141 & 215 & 1.525 (2.64) & 57 = 40.4\% [32.6, 48.9] \\
Scribe C & 142 & 142 & 275 & 1.937 (3.26) & 60 = 42.3\% [33.8, 50.0] \\
\midrule
\textbf{Pooled} & \textbf{565} & \textbf{142} & \textbf{618} & \textbf{1.094 (2.32)} & \textbf{177 = 31.3\% [27.0, 35.6]} \\
\bottomrule
\end{tabular}
\caption{\textbf{The census: 618 verified failures in 565 notes from three deployed scribes.} Three commercial ambient scribes wrote notes from the same consultations. Every note went through 12 separate discovery passes, which produced 13,678 candidate findings. The 5,898 the discovering model rated high or medium importance went to a panel of two skeptics from different model families, \texttt{anthropic/claude-opus-5} (the harsher) and \texttt{openai/gpt-5.5} (the gentler), with splits decided by \texttt{openai/gpt-5.4}. That tiebreak model shares a developer with the gentler skeptic, and Section 4.2 records which way every tiebreak went. The panel verified 618, 10.48\% [8.9, 12.1] of the candidates put to it. Findings per note is verified findings divided by that product's notes, with the standard deviation beside it; over all 565 notes the median is 0 and the maximum 18. Intervals on the notes-with-at-least-one column and the candidate-survival figure are 95\% bootstrap intervals resampling whole consultations, since notes from one consultation share a transcript (Section 3.1). Scribe A is generated through an API at two note templates per consultation, so it contributes two notes where the other two contribute one, and its two templates land almost identically (0.447 and 0.461 findings per note), so the gap to the other products is not a property of one template. Notes span four strata: 228 PriMock57, 180 ACI-Bench, 120 authored, 37 trap-blind.}
\label{tab:census}
\end{table*}

The products differ - Scribe A 0.454 verified findings per note, Scribe B 1.525, Scribe C 1.937 - and we do not read that spread as a product ranking, for three reasons. How each product received the consultation differs: audio replayed to a listening product for B and C, an API for A. Scribe A contributes two notes per consultation where the others contribute one, though its two templates behave almost identically (Table~\ref{tab:census}), so the gap is not a property of one template. And the spread depends on the review standard: re-reviewed under a lenient instruction (Section 4), the gap between products in the share of candidates verified narrows from four-and-a-half-fold to 1.2-fold and Scribes B and C change order, though Scribe A reads lowest under both standards. The pooled rate is shaped by Scribe A's doubled output too: it supplies half the notes at the lowest rate, so pooling pulls the headline down rather than up. What the spread does establish is that the pooled rate is not one bad product's doing: notes with at least one verified failure run 21.3\% {[}16.0, 27.0{]}, 40.4\% {[}32.6, 48.9{]} and 42.3\% {[}33.8, 50.0{]} across the three.

Sorting the 618 findings needed a scheme, and the top of it was fixed before clustering. The top tier is the four-class split the audit literature uses - omission, addition, wrong output, irrelevant or misplaced text \citep{hose2025development} - and the second tier is the mechanism: thirteen categories, eleven of them hunted by a targeted pass (the two others were not hunted; Appendix A.8). At the top tier the panel's 618 verified findings divide into wrong output 207 (33.5\%), addition 181 (29.3\%), omission 143 (23.1\%), irrelevant or misplaced text 46 (7.4\%), and 41 (6.6\%) unmapped - open-pass findings whose descriptions matched none of the matcher's vocabulary, so the scheme under-places them rather than misplacing them. Against the written clinical rubric, 476 of the 617 graded findings are critical - an error that would plausibly change clinical action or safety - 140 supporting and 1 peripheral. That distribution describes the survivors, not scribe output at large: the discovery filter and the panel each remove low-consequence material before a finding reaches the rubric, and which way the rubric leans is rater-dependent across three blinded gradings (Appendix A.1).

\begin{figure*}[tp]
\centering
\includegraphics[width=\linewidth]{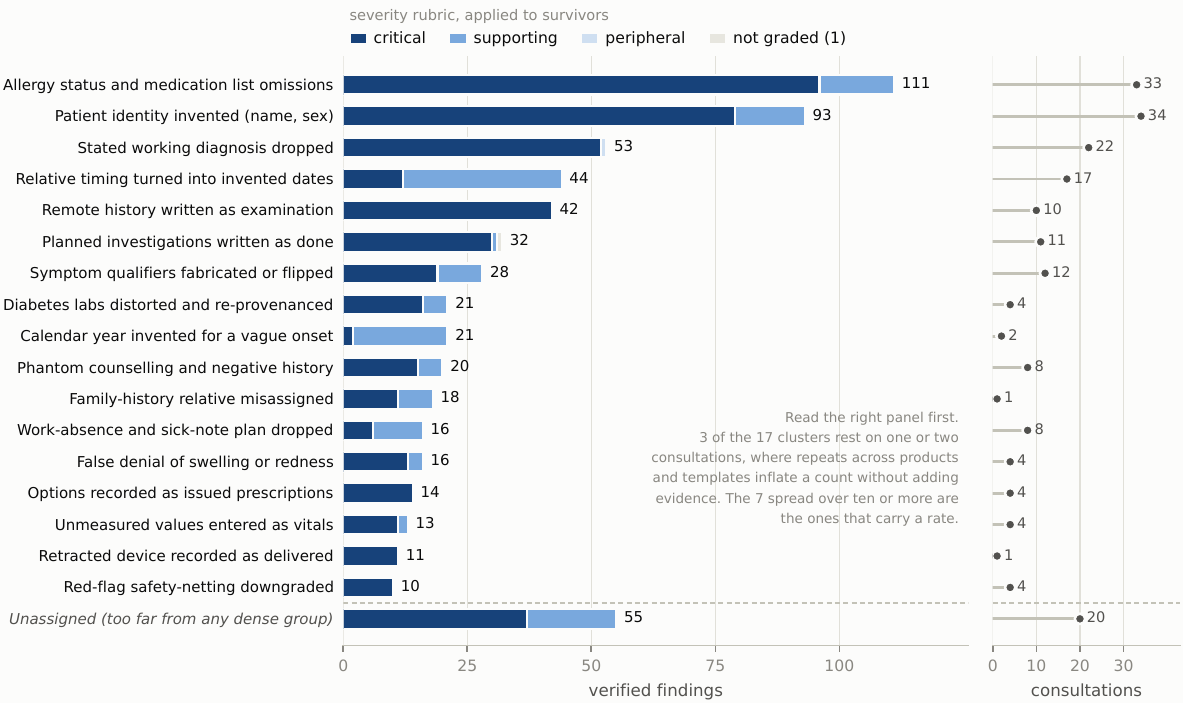}
\caption{\textbf{The seventeen subcategories the census discovered, ranked by size.} The seventeen row labels are shortened forms of the subcategory names in Table~\ref{tab:clusters}, in the same order. The labels are the clustering step's own, and the largest group's failures run in both directions - 20 of its 111 members came from the fabrication hunt - so its omissions label names the group, not every member (Section 3.3). Bar length is verified findings, split by the severity rubric applied to survivors: critical, an error that would plausibly change clinical action or safety; supporting, one that degrades the record without changing what happens next; peripheral, no plausible consequence. One finding carries no grade. The rubric reproduces to within a grade across three blinded gradings, which do not agree on which way it leans, so the critical shares should be read as rater-dependent (Appendix A.1). The right panel is the number of distinct consultations each group's findings came from, and it should be read first. The 55 findings the clustering left unassigned are shown below the rule rather than folded into a neighbouring group.}
\label{fig:ladder}
\end{figure*}

Only the third tier is discovered from the data: the verified findings' descriptions are embedded and clustered by density, giving 17 subcategories covering 563 of the 618 findings; re-clustering under different seeds gives 17 to 21 groups with largely the same boundaries (mean adjusted Rand index 0.727, where 1.0 would be identical groupings and 0 chance agreement), coarser or finer clustering parameters give 17 to 54, and each cluster's place in the scheme was assigned by the labelling call (recipe, parameter sweep and stability in Appendix A.2). All 17 are ranked, with their severity mix and the number of consultations each rests on, in Figure~\ref{fig:ladder}; the five largest are described below, and all 17 are listed with examples in Table~\ref{tab:clusters}. Product columns in Table~\ref{tab:clusters} are counts of findings, not rates - Scribe A writes two notes per consultation - so they cannot be compared across products without Table~\ref{tab:census}'s denominators; cluster sizes likewise count verified findings rather than distinct errors, and Appendix A.9 gives the deduplicated count for six of them.

{\footnotesize\setlength{\tabcolsep}{3pt}\renewcommand{\arraystretch}{1.05}
\begin{longtable}{@{}>{\raggedright\arraybackslash}p{0.19\linewidth}rrrrrr>{\raggedright\arraybackslash}p{0.10\linewidth}>{\raggedright\arraybackslash}p{0.30\linewidth}@{}}
\caption{\textbf{All 17 subcategories, from the 618 verified findings.} The first two tiers of the taxonomy were fixed before the findings were sorted; the third is discovered from the data. We embed each verified finding's description, project it with UMAP and cluster by density with HDBSCAN, then label each cluster in one call (full recipe and parameters in Appendix A). 563 of the 618 verified findings fall in a cluster and 55 are left unassigned. The count carries measured instability: re-running the same recipe under UMAP seeds 42, 43 and 44 gives 17, 20 and 21 clusters with largely the same boundaries (pairwise adjusted Rand index 0.667, 0.707 and 0.808, mean 0.727, where 1.0 is identical groupings), so the count is stable only to within a few. Columns A, B and C count findings rather than notes; high importance is the verification panel's own rating, out of the cluster's n. Placement in the fixed top two tiers is one call for the whole cluster, so cluster sizes do not sum to the top-level class totals in Section 3.2; 16 of the 17 clusters place under a tier-2 category and one has no category in our scheme, whose classes come from the published taxonomies. Examples are paraphrased from our own finding descriptions rather than quoted from the notes. Tier-1 and tier-2 names are plain-language forms of the released scheme's category identifiers; Table~\ref{tab:checkability} carries each hunted category's description.}\label{tab:clusters}\\
\toprule
& & \multicolumn{3}{c}{product} & & high & place in &\\
\cmidrule(lr){3-5}
Subcategory & n & A & B & C & consultations & importance & the scheme & example\\
\midrule
\endfirsthead
\toprule
& & \multicolumn{3}{c}{product} & & high & place in &\\
\cmidrule(lr){3-5}
Subcategory & n & A & B & C & consultations & importance & the scheme & example\\
\midrule
\endhead
\midrule
\multicolumn{9}{r}{\footnotesize\emph{continued on the next page}}\\
\endfoot
\bottomrule
\endlastfoot
Allergy status and medication list omissions & 111 & 19 & 40 & 52 & 33 & 71 & omission & Allergy status was asked and answered in the consultation, and the note documents none while recording a prescription. (Scribe B) \\
Invented patient identity: name and sex & 93 & 10 & 45 & 38 & 34 & 49 & addition / fabrication & The patient's name is inaudible in the transcript, and the note supplies one that is a chemotherapy brand name. (Scribe C) \\
Stated working diagnosis dropped from note & 53 & 53 & 0 & 0 & 22 & 52 & omission & The clinician tells the patient the symptoms suggest gastroenteritis; the note carries the advice that follows from it and no impression at all. (Scribe A) \\
Relative timing converted to invented calendar dates & 44 & 0 & 4 & 40 & 17 & 2 & wrong output / timing & The transcript says 'yesterday' and the note prints a specific calendar date. (Scribe C) \\
Remote consult history written as objective exam & 42 & 0 & 11 & 31 & 10 & 33 & addition / examination provenance & A telephone consultation in which no examination took place, written up with examination findings as observed. (Scribe C) \\
Planned investigations documented as already done & 32 & 4 & 15 & 13 & 11 & 21 & wrong output / modality hardening & The transcript has an ECG still to be arranged with reception; the note records it as performed in clinic today. (Scribe C) \\
Fabricated/flipped symptom qualifiers (sleep, timing, location) & 28 & 0 & 14 & 14 & 12 & 11 & addition / fabrication & The note reports the symptoms as affecting sleep, which the consultation never mentions. (Scribe B) \\
Diabetes lab values distorted and re-provenanced & 21 & 6 & 11 & 4 & 4 & 6 & wrong output / dose value & The transcript gives a recalled HbA1c of about 7.7 per cent and the note records 7.0 per cent. (Scribe B) \\
Fabricated calendar year for vague onset date & 21 & 0 & 12 & 9 & 2 & 0 & wrong output / timing & 'Last December' becomes December 2025, a year the consultation never states and the patient's stated age contradicts. (Scribe C) \\
Phantom counselling and unasked negative history & 20 & 0 & 11 & 9 & 8 & 4 & addition / fabrication & The clinician asks whether a previous doctor discussed the risks of aspirin; the note records that the advice was given in this consultation. (Scribe C) \\
Family-history relative and lineage misassignment & 18 & 0 & 0 & 18 & 1 & 4 & wrong output / attribution & A cancer the transcript places in the maternal grandmother is recorded against the patient's mother. (Scribe C) \\
Dropped work-absence and sick-note plan elements & 16 & 11 & 5 & 0 & 8 & 3 & omission & The clinician tells the patient not to go into work until this is sorted out, and the plan omits it. (Scribe A) \\
False denials of swelling/redness & 16 & 0 & 6 & 10 & 4 & 7 & wrong output / negation & The patient reports inflamed knees and the examination finds swelling and redness; the note records both as denied. (Scribe C) \\
Alternative options recorded as issued prescriptions/orders & 14 & 6 & 0 & 8 & 4 & 13 & wrong output / modality hardening & A choice offered between two antibiotics is listed as two issued prescriptions. (Scribe C) \\
Unmeasured or corrupted values entered as objective vitals & 13 & 0 & 1 & 12 & 4 & 7 & addition / examination provenance & A temperature the patient took at home enters the vitals as an observed measurement, its value corrupted to an implausible 13.7\textdegree{}C. (Scribe C) \\
Retracted device captured as delivered care & 11 & 7 & 4 & 0 & 1 & 5 & no published category & The clinician retracts the thumb spica in the next breath and substitutes a wrist brace; the note records the spica as applied. (Scribe A) \\
Red-flag safety-netting downgraded to routine follow-up & 10 & 6 & 1 & 3 & 4 & 7 & wrong output / negation & Sudden weight gain is given as a red flag warranting urgent contact, and the note attaches it to the routine booking instruction. (Scribe A) \\

\end{longtable}}

\subsection{The two largest clusters: allergy and medication state, and invented identity}

The largest cluster is \textbf{allergy status and medication-list failures} (n=111, about 47 distinct errors, over 33 consultations; 71 rated high importance by the panel; the product split is in Table~\ref{tab:clusters}) - one of two top-five clusters with findings from all three products. Its labels in Table~\ref{tab:clusters} and Figure~\ref{fig:ladder} are the clustering step's own; the failures run in both directions, and 20 of its 111 members came from the fabrication hunt. Verified cases include a prescribed dose understated in the note: the transcript specifies up to two 500mg paracetamol tablets four times a day, ``that's the full dose''; the note carries ``500mg up to four times daily'' - half the dose, asserted as the regimen (Scribe B). A medication the patient actually takes for menstrual cramps, ibuprofen, is replaced in the medication list by nifedipine, a blood-pressure drug the consultation never mentions (Scribe C). A prescribed emollient, Cetraben, becomes cetirizine, an oral antihistamine (Scribe C). One member runs the other way, and the panel and the rubric both graded it at the top: in a chest-pain consultation, the patient's answer to the allergy question is inaudible on the recording, and the note records a definite aspirin allergy - a fabrication that could wrongly withhold aspirin in suspected acute coronary syndrome (Scribe B). Both directions produce the same defect: the note asserts an allergy and medication state the consultation does not support, whether by dropping what was said or by recording what was not.

The second largest, and the second of the two top-five clusters with findings from all three products, is \textbf{invented patient identity} (n=93, about 49 distinct errors, over 34 consultations; 49 high importance). Where a name is inaudible or a sex never stated, the note supplies one and uses it throughout: a patient who gives his name as John Smith is documented as ``a 32-year-old woman'' (Scribe C); several notes gender an unspecified patient female from the first line (Scribe A). In one consultation where the patient's name is inaudible, the note names them ``Gemzar'' - a chemotherapy brand name that appears nowhere in the consultation (Scribe C). No product here had a patient record to draw on, and an integrated deployment that prefills demographics would not need to infer them. What these findings measure - and what transfers to deployment wherever the audio leaves a field unfilled - is that the products fill unknowns with invented specifics rather than flagging them. That is the shape of every verified identity finding, not a measured rate: we did not count how often a product flagged an unknown instead. What a product does when a prefilled record and the audio disagree is a different question, and no product here was given a record to disagree with. This cluster and the invented-date cluster below are the two the design most plausibly inflates, so Appendix A.7 reports the census with each removed in turn while every note stays in the denominator: the headline falls from 31.3\% {[}27.0, 35.6{]} to 24.8\% {[}20.8, 29.0{]} with both classes out, and which product reads worst inverts on the way down.

\subsection{The next three: dropped diagnoses, invented dates, invented examinations}

\textbf{A stated working diagnosis dropped from the note} (n=53, about 32 distinct errors; Scribe A only; 22 consultations; 52 of 53 high importance): the clinician tells the patient ``I think your symptoms are suggestive of something called gastroenteritis'', and the note carries the advice that follows from the diagnosis and no impression at all, leaving the plan without a documented rationale (Scribe A). This is a single-product result, present in both of that product's note templates (30 and 23 findings), so it reads as a property of that product's note format rather than of one template. The dropped-diagnoses family in the replication count below is wider than this cluster - its expression fires on any description naming a diagnosis, an impression or an assessment - so pairings recorded there are the matcher's families, not this cluster.

\textbf{Relative timing converted to invented calendar dates} (n=44, about 20 distinct errors; Scribes B and C; 17 consultations): the patient says the pain ``started yesterday'' and the note prints ``beginning mid-morning on August 9, 2026'' (Scribe C). A deployed scribe legitimately knows the encounter date and may resolve ``yesterday'' against it; when we ran these consultations through the products they resolved against the day we ran them, and what the findings record is a resolved date printed as stated fact - not whether the arithmetic behind it was right - in notes that elsewhere invent the specifics outright (a calendar year for a vague onset, contradicting the patient's own stated age, is its own cluster in Table~\ref{tab:clusters}). The panel rated almost none of it important - 2 of 44, though the clinical rubric grades 12 of the 44 critical; the two axes are defined in Appendix A.1. An invented specific is a fluent failure, plausible on its face and, we would expect, easy to read past - an inference from the form of these findings, since nothing here measures what a signing clinician catches (Section 6). Of the two classes a patient record might have prevented, this is the weaker case, since the encounter date has a legitimate source in deployment and an invented name has none. Appendix A.7 therefore removes the dates at a separate step from the identities.

\textbf{Remote-consultation history written as objective examination} (n=42, about 13 distinct errors; Scribes B and C; 10 consultations; 33 high importance): on telephone consultations where no examination could have occurred, both products document examination findings as observed - the opening case, now with its numbers. The ten are the consultations that produced these findings, in each of which the transcript itself establishes that no examination occurred. The corpus carries no reliable label for which of its other consultations were remote, so how many examination-free consultations produced no such finding is unknown, and we state this as a count, not a rate.

The same failure types recur across products. A fixed keyword matcher groups findings into eleven coarse failure families - a separate grouping that cuts across the thirteen mechanism categories above, covering examination-not-performed, invented demographics, timing errors, dropped diagnoses and seven more. It runs over the failure-mode label the discovery pass gave each finding plus its free-text description, and leaves anything it cannot place under that label (the same deterministic matcher that places open-pass findings; Appendix A.9). 330 consultation-and-family pairings hold at least one verified finding, and in 45 of them (13.6\%) two or more products produced a failure of the same family independently - 31 consultations, 30 of the 45 rated high importance. That figure has no chance baseline to sit against, since a product can only replicate a family that a discovery pass proposed for its own note, so we read it as an existence result rather than a rate: nine of the eleven families replicate this way within a single consultation, and the remaining two - laterality and site, and negation - appear on more than one product but never on the same consultation, so no family in this census is one vendor's defect alone. These families are not the clusters of Table~\ref{tab:clusters}, and a name shared between the two does not make them one group: two products can land in the same family on findings that sit in different clusters, or in none, so the dropped-diagnosis cluster above stays Scribe A's alone while the dropped-diagnosis family does not. The matcher is deliberately loose, so this is replication of failure types, not of individual errors, and which product fails on which consultation is not settled by it. An NHS trust evaluation of four products in acute services reports the same shape from live use: accuracy varied with the clinical workflow rather than with the product chosen \citep{thehill2026avt}.

\subsection{A failure mode our scheme could not place}

One cluster could not be placed under any category of our scheme, whose classes are drawn from the published scribe-error taxonomies: \textbf{a retracted device recorded as delivered care} (n=11; Scribes A and B). The clinician proposes a thumb spica, retracts it in the next breath - ``sorry yeah not a thumb spica we're gonna brace your arm'' - and the note records ``Thumb spica applied'' (Scribe A), sometimes clashing with its own plan section. The failure is neither invention (the device was spoken) nor a negation error: it is treating an explicitly retracted utterance as the final decision. All 11 findings come from one consultation across two products - about two distinct errors once passes are deduplicated (Appendix A.9) - so we name it as a failure mode rather than a rate. The placement is one labelling call's: the member-vote rule that would have overridden it never reached its threshold on any cluster, and no human adjudicated the declination (Appendix A.2), so the argument for treating this as a class of its own is the one made here, not the call.

\section{What a failure rate means: the instrument, and the published audits}

The published scribe audits disagree with each other by a margin an instrument difference alone can produce. Omission is 54\% of errors in one of Biro's products and 83\% in the other, 76.3\% in Anderson, 86.3\% in Kernberg; our census finds 23.1\% of verified findings. Products differ too, and the published spread is not all instrument: Biro's two, read by the same reviewers under one standard, sit 29 points apart on their own, most of the 32 points that separate the highest published share from the lowest. None of those three studies reports what its review standard contributed to its numbers, so the published record cannot say how much of the spread is scribe and how much is standard. Across studies these figures are not comparable as published: the review standard differs every time, and sometimes the denominator does too. (Taylor's pilot counts on a different denominator, the share of notes rather than the share of errors; the like-for-like comparison with it closes this section.) We can report what our standard contributed, because we measured it: a sample of our own candidates, re-reviewed with everything held fixed except the review instruction.

\subsection{The same candidates under four review standards: the instruction moves the count eight and a half times over, the machinery one point}

The panel's instruction tells each skeptic to refute a candidate if refuting it is at all defensible and to default to refuting when uncertain. To measure what that instruction contributes to the census, we drew a stratified fifth of it: 115 of the 565 notes, drawn as whole notes, carrying 1,295 candidates across 87 consultations. One drawn note carries no candidate, and it stays in the note-level denominator. We then re-reviewed every drawn candidate under a lenient instruction in place of the panel's. The lenient instruction asks the reviewer to apply the standard an informal reviewer would use before signing the note off, and to keep a finding if a reasonable reviewer would want it corrected before the note enters the record, even where it is minor, arguable, or a matter of documentation practice rather than clinical fact.

Everything else was held fixed: the same model as the panel's harsher skeptic, the same note and transcript evidence assembled by the same code, the same transport and settings, with a runtime check that fails the run if the two prompts differ anywhere beyond the review instruction and its closing question (both are printed in Appendix A.3). The census run had also stored each skeptic's solo verdict on every candidate before any tiebreak, so the same sample supports four review standards side by side: each strict skeptic alone, the full panel, and the lenient re-review (Table~\ref{tab:standards}).

% T5 - the same candidates under four review standards. Generated by
% extract_p2_floats.py from results/second-panel/; do not edit by hand, regenerate.
\begin{table}[tp]
\centering
\footnotesize
\setlength{\tabcolsep}{3.5pt}
\begin{tabular}{@{}lrrccc@{}}
\toprule
Review standard & verified & share verified & 95\% interval & notes flagged & omission share\\
\midrule
Harsher-family skeptic alone, strict instruction & 121 & 9.3\% & [6.0, 12.9] & 27.8\% & 22.3\%\\
Gentler-family skeptic alone, strict instruction & 230 & 17.8\% & [13.1, 22.5] & 54.8\% & 33.9\%\\
Full panel: both skeptics and the tiebreak & 134 & \textbf{10.3}\% & [6.4, 14.6] & 27.8\% & 23.9\%\\
Harsher skeptic's model, lenient instruction & 1,023 & \textbf{79.0}\% & [75.8, 82.0] & 96.5\% & 28.2\%\\
\bottomrule
\end{tabular}
\caption{\textbf{The same 1,295 candidates under four review standards: the instruction sets the count.} A stratified fifth of the census (115 of the 565 notes, drawn as whole notes, 87 consultations, 1,295 candidates). The census run had recorded each skeptic's solo verdict on every candidate before any tiebreak, so the first three standards are re-read from stored verdicts; the fourth is one new call per candidate on the same model as the harsher skeptic, with the evidence, transport and settings unchanged and only the review instruction replaced (both instructions are printed in Appendix~A.3). Bracketed intervals are 95\% cluster bootstrap intervals over consultations. The notes-flagged column is the share of the 115 sampled notes carrying at least one verified finding under that standard; the census panel's 27.8\% here is consistent with its 31.3\% [27.0, 35.6] over all 565 notes. The first and third rows share that cell only by offsetting exchanges: against the harsher skeptic alone the panel keeps 38 candidates and loses 25, and at note level five sampled notes leave the flagged set while five others enter it. The omission column is the share of that standard's verified findings placed under omission at tier 1 - the class mix moves with the standard on identical candidates (Section~4.4). The strict and lenient standards are strictly nested: the lenient review keeps all 134 of the panel's survivors and adds 889, and no candidate goes the other way.}
\label{tab:standards}
\end{table}

The instruction is worth a factor of eight and a half, or - less sensitive to where the candidate pool starts - 69.7 percentage points. The same model, shown the same evidence, verifies 121 of the 1,295 candidates when told to refute, 9.3\% {[}6.0, 12.9{]}, and 1,023 when told to review leniently, 79.0\% {[}75.8, 82.0{]}, a difference of 69.7 points {[}65.5, 73.6{]}. Candidate-level intervals in this subsection are cluster bootstrap intervals over the 87 consultations, because candidates from one consultation share a transcript; the note-level shares below carry Wilson intervals over the 115 sampled notes, and are labelled so in Appendix A.3. On the paired candidates the movement is entirely one-directional: 902 candidates change from refuted to verified, and none changes the other way.

The two standards are strictly nested. The lenient review keeps every one of the 134 candidates the full panel had verified on this sample and adds 889; no candidate the panel kept is refused by the lenient review. The two standards therefore sit at opposite ends of one axis, and the pair reads as a measured range rather than a bound: the same products, on the same material, support a verified count anywhere from the 134 findings the panel kept to 7.6 times that number {[}5.4, 12.4{]}, depending on the standard applied.

The panel's machinery is not what moves the count. At the strict instruction, going from the harsher skeptic alone to the full machinery - the second model family plus the tiebreak - moves the rate from 9.3\% to 10.3\%. That is one percentage point, clustered interval {[}-1.2, +3.7{]} points, and the paired counts (38 candidates gained, 25 lost) do not distinguish it from no change (p = 0.13). The panel gains candidates over its harsher member because a split goes to the tiebreak, which upholds some of what that member alone refused, so it is not a conjunction of its two skeptics. The choice of model family matters more: at the same strict instruction the gentler family alone verifies 17.8\% {[}13.1, 22.5{]} against 9.3\% for the harsher family, a gap of 8.4 points {[}5.0, 12.1{]} - larger than the machinery's effect, and still roughly a factor of eight below the 69.7 points the instruction moved it. What the two-family design provides is different: cross-family agreement on 86.2\% of the census's 5,898 candidates, a recorded dissent and a genuine third opinion on the 812 splits among them, and the decomposition in this paragraph, which exists because every solo verdict was stored.

The same experiment puts the same range around the census's headline. Under the panel, 27.8\% {[}20.5, 36.6{]} of the 115 sampled notes carry at least one verified failure - consistent with the 31.3\% {[}27.0, 35.6{]} the census reads over all 565 notes, which is the check that the sample reproduces the population. The harsher skeptic alone flags the same 27.8\% of notes, and the gentler skeptic alone flags twice as many, 54.8\% {[}45.7, 63.6{]}. Under the lenient instruction almost every note is flagged, 96.5\% {[}91.4, 98.6{]}, at 8.9 verified findings per note against 1.2 under the panel (Table~\ref{tab:standards}). The census's own headline is therefore anchored to one reviewer: on these notes an equally strict reviewer from the other model family reads roughly twice the note-level rate. Under a standard resembling an informal read before sign-off, essentially every note in the sample has something a reviewer would want corrected; under an adversarial standard, fewer than a third do.

Neither figure is the true rate. No external ground truth adjudicates between the standards: the physician review in Section 4.3 calibrates the strict panel's refusals, and nothing here disturbs it. The lenient figure has cautions of its own. It is one call from one model family, and since the gentler family verifies more at the strict instruction, 79.0\% is better read as a floor for lenient review than a ceiling. A ratio measured from a base of 9.3\% cannot exceed about eleven-fold whatever the lenient standard does, so the 8.5x figure is partly a property of how low the strict rate is; the 69.7-point difference, and the fact that the lenient standard keeps every finding the panel kept, are the more stable statements. What the experiment fixes is the size of the standard's contribution to the count, not which count is right.

Our own records had pointed the same way with less control. An earlier version of this study's panel - three skeptics from a single model family, an earlier model generation and a 121-note pilot corpus - verified 37.3\% of the candidates put to it, against 10.48\% {[}8.9, 12.1{]} of the 5,898 candidates the final instrument reviewed. Restricted to the candidates the final instrument's importance filter would have kept, the pilot panel's rate is 41.0\%, so the like-for-like contrast is slightly wider still. The model generation, the family mix, the discovery depth and the corpus all moved at once, and Appendix A.3 sets those configurations out row by row and says which changes are confounded in each step. Two concurrent results, both from vendor research teams and each on its own product, show the same dependence under control: relabelling clinically reasonable inference as supported moves the flagged rate on the same SOAP notes from 35.2\% to 9.1\% \citep{vachhani2026beyond}, and a vendor preprint reports its error burden under two detectors of deliberately different sensitivity, 24.4\% against 6.2\% on the same notes \citep{bergman2026quality}. To our knowledge ours is the only such reading that is independent of any vendor, covers more than one product, and decomposes the effect - instruction, model family, panel machinery - with the two standards strictly nested.

The blinded sittings of Section 3.1 add a human reading of the same effect, now from two clinicians rather than one: shown high-importance candidates the panel had refused, blind among verified findings, both judged every one a genuine failure - nine of nine to the physician author and four of four to the independent clinician. The stratum they were drawn from holds 1,697 candidates the panel rated important and refused, so thirteen items is a small and uneven basis (the four span three consultations, and two of them are the same medication exchange written up by two products), but on the items they saw, both clinicians applied a standard closer to the lenient instruction's than to the panel's, and the panel's count is the strict end of the range.

\subsection{What survives verification is what can be checked}

The panel does not verify all kinds of claim at the same rate, and the ordering says what it is actually doing (Table~\ref{tab:checkability}). The three categories it verifies most often are the ones a skeptic can settle by looking at the text: a dose is a number that is right or wrong (32.2\% of candidates verified), a fabricated fact either appears in the transcript or does not (24.6\%), an attribution names a person the transcript names or does not. The three it verifies least often all require a judgement about what should have been in the note, or how firm a claim was entitled to be: plans and possibilities written as done and definite, text under the wrong heading, and facts missing from the note - omission, at 6.1\%, which is not the lowest of the three (each category's rate is in Table~\ref{tab:checkability}). Omission is the one a reader will look for, but all three of the categories that need a normative judgement get verified at a fifth to a seventh of the top category's rate. These shares are of candidates that had already cleared the discovering model's own importance filter, and that filter's pass rate by category is not reported here, so part of the gradient may sit upstream of the panel.

% T3 - survival by checkability. Generated by extract_p2_floats.py from
% master/findings_verified_master.json; do not edit by hand, regenerate.
\begin{table}[tp]
\centering
\small
\setlength{\tabcolsep}{5pt}
\begin{tabular}{@{}lrrr@{}}
\toprule
Category & candidates & verified & share verified\\
\midrule
Dose or measured value wrong & 115 & 37 & 32.2\% \\
Fabricated content & 395 & 97 & 24.6\% \\
Wrong person or source attributed & 135 & 28 & 20.7\% \\
Polarity flipped (denial for report, or reverse) & 167 & 24 & 14.4\% \\
Open discovery pass (no assigned category) & 1,396 & 196 & 14.0\% \\
Left and right confused & 35 & 4 & 11.4\% \\
Timing wrong or invented & 366 & 39 & 10.7\% \\
Examination provenance (history written as examined) & 225 & 18 & 8.0\% \\
Note contradicts itself & 341 & 26 & 7.6\% \\
Omission (fact absent from the note) & 1,284 & 78 & 6.1\% \\
Text under the wrong heading & 887 & 46 & 5.2\% \\
Plans or possibilities written as done or definite & 552 & 25 & 4.5\% \\
\midrule
\textbf{All categories} & \textbf{5,898} & \textbf{618} & \textbf{10.48\%} \\
\bottomrule
\end{tabular}
\caption{\textbf{What survives adversarial verification is what can be checked.} Candidate findings per discovery category, the number the two-family verification panel verified, and the share verified. The three categories verified most often are the ones a skeptic can settle by looking at the text - a dose is right or wrong, a fabricated fact appears in the transcript or does not, an attribution names a person the transcript names or does not. The three lowest all require a judgement about what should have been in the note or how firm a claim was entitled to be - plans or possibilities written as done or definite, text under the wrong heading, and omission - and all three are verified at a fifth to a seventh of the top category's rate. Omission is not the lowest of them. Categories are the eleven targeted discovery passes plus the open pass; the open pass's survivors are distributed into the published classes before any comparison with the literature (Appendix~A.9).}
\label{tab:checkability}
\end{table}

The gradient shows what adversarial verification gains, and what it costs. The panel does not wave findings through: it refutes roughly nine candidates in ten, and the harsher skeptic refutes 90.6\% of everything it sees. The two families agree unaided on 86.2\% of candidates - Cohen's kappa 0.44 once chance agreement between two mostly-refusing reviewers is subtracted, which is moderate, and is why a lone dissent forces a third opinion rather than killing a finding by itself. The tiebreak model shares a developer with the gentler skeptic, so the run records which way every one of its 812 tiebreak calls went: it sided with the opposing family's skeptic 531 times and its own 281, upholding 195 findings and cutting 617. The cost is that a standard favouring checkability holds the judgement-call classes to a demanding bar - which is the direction a floor should lean, and the direction to keep in mind when our omission share meets the published audits' below.

The gradient is itself partly a property of the strict standard. Re-reviewed under the lenient instruction of Section 4.1, the ordering survives in outline: across the eleven categories with at least twenty sampled candidates, the lenient rates put the categories in much the same order as the panel's (0.72 on a rank correlation, where 1.0 would be the identical order), with the checkable categories still at the top and the judgement categories still at the bottom. But the ordering compresses from an eleven-fold spread (2.8\% to 30.8\% of sampled candidates verified) to a spread of one and a half (63.0\% to 94.3\%): a lenient reviewer verifies most of nearly everything, so the ordering persists while ceasing to discriminate. One category changes places outright. The panel verifies text filed under the wrong heading at 4.3\% on this sample (5.2\% across the 887 such candidates in the whole census); the lenient reviewer verifies it at 88.5\%. Misplaced text is plainly worth fixing when the question is whether a reviewer would want it fixed, and almost always refutable when the instruction is to refute if at all defensible. The same compression appears on the panel's own importance ratings (a 3.2-fold spread across high, medium and low importance becomes 1.2-fold) and on the between-product comparison (Section 3.2). So the ordering in Table~\ref{tab:checkability} describes what survives adversarial review; for at least one category, that is a property of the review standard rather than of the notes.

\subsection{We audited our own refusals: most are judgement calls, and a quarter of the sample turned on restatement}

Omission is the largest of the eleven targeted hunts - 1,284 of the 5,898 candidates, second only to the open pass's 1,396 - and the panel verified just 6.1\% of it. A near-zero verified omission rate could mean the scribes rarely omit important material or that the instrument suppresses omissions, and the rate alone cannot say which. So we audited the refusals as part of the run. A separate model pass re-read a stratified sample of 40 refused omission candidates, this time with the full note in view, and classified why each was refused. That auditor is the gentler skeptic, \texttt{openai/gpt-5.5}, in a second job, so on every unanimous refusal it re-read a verdict it had itself cast, though never one on a candidate it had generated.

Two thirds of the refusals were importance or scope calls a review panel is entitled to make (42.5\% not material, 22.5\% not required in a note of this kind), and in one sampled case (2.5\%) the candidate itself was spurious - the transcript never contained the fact it claimed, so refusing it was right. The remaining third is where the instrument is doing something a reader should know about. 7.5\% were judged wrongly cut - and the blinded physician review of ten refusals endorsed the panel on every one they could judge, including all three the model auditor had flagged, so even that 7.5\% reads as a model-generated upper bound rather than a measured rate of real omissions wrongly cut. We treat that review as calibration rather than gold standard, and concurrent work says why: when ten clinicians were asked whether flagged documentation errors were genuine, they agreed with each other only fairly (Gwet's AC1 0.24, a chance-corrected agreement coefficient read on the same scale as kappa), leaving no human consensus to serve as truth \citep{bergman2026judges}.

A second check runs the same direction: among omission candidates the panel itself rated high-importance, 15.3\% were verified - essentially the same as the 15.7\% verified among high-importance candidates of every kind. That comparison conditions on the panel's own importance rating, so it cannot see the panel's behaviour at the rating stage, where it calls 26.5\% of omission candidates high importance (340 of 1,284) against 34.1\% of candidates overall (2,014 of 5,898); what it establishes is that once an omission is rated important the panel treats it like anything else. The low overall omission share is therefore the panel making importance judgements on medium- and low-importance material, not a bar applied differently to omissions the panel itself rated important; the largest clusters other than the invented dates are dominated by high-importance findings and stand on the panel's strictest ground.

The remaining quarter is a different kind of refusal. In 10 of the 40 sampled refusals - 25.0\% (Wilson {[}14.2, 40.2{]} over those 40) - the panel refused an omission candidate because the fact is stated somewhere else in the note. Clinical notes repeat themselves - duplication across successive notes runs at 54 to 78\% in hospital records \citep{wrenn2010quantifying}, and the property we meet here is its within-note form. A verification standard that counts a restatement anywhere as capture will refuse omission candidates that a clinician reading the right section would count - a fact recorded outside the section where a reader would look for it is the defect the published taxonomies call misplaced text. That 25\% share is measured on 40 sampled refusals and moves with the sample - a 15-refusal pilot of the same audit had put it at 0\% - and the blinded physician review leaves this bucket essentially unvalidated: eight of its ten items printed only the part of the note covering the point, and of the three present-elsewhere items in the pack the physician answered one and abstained on the other two (Appendix A.4).

The scope choice behind this is made both ways in print and its effect on a count has not been measured. One widely used annotation scheme marks an omission against a required location in the note, so a fact recorded in the wrong section counts as missing from the right one \citep{benabacha2023investigation}; a vendor-co-authored checklist scopes itself to one section at a time and drops a cross-section completeness question for exactly that reason \citep{zhou-etal-2025-feedback}; MED-OMIT likewise checks omissions against a single note section; our panel sits at the opposite pole and counts a restatement anywhere as capture. We know of no published audit that has measured what that choice does to its count, which is what the 25\% above begins to do; we publish it beside our omission rate as part of the instrument, and the companion paper finds the same redundancy of clinical notes defeating automated judges too.

\subsection{Reading the published audits with the instrument in hand}

\begin{figure*}[tp]
\centering
\includegraphics[width=\linewidth]{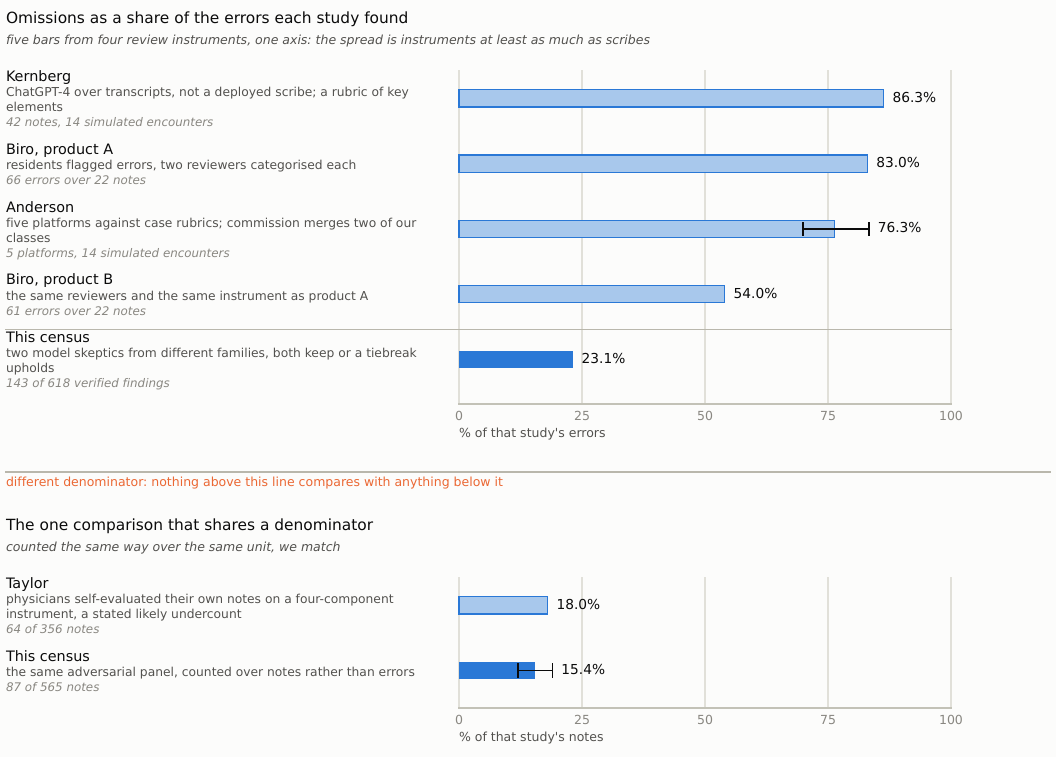}
\caption{\textbf{The published omission shares Section 4 reconciles, each beside the review standard that produced it.} Above the rule, omissions as a share of the errors each study found, with how those errors were identified and categorised and what the share is computed over. The five bars above the rule come from four review instruments, because Biro's two products were read under one; Taylor's, below the rule, is a fifth. The five instruments, one per study, were these: residents flagging errors in notes they had just read aloud, with two reviewers categorising each and disagreements settled by consensus (Biro, both products, one instrument); two clinical experts scoring against a rubric of key reportable elements built from the transcript and the case storyboard (Kernberg, which prompted ChatGPT-4 over transcripts rather than auditing a deployed scribe); five platforms graded against case rubrics with a commission class that merges what we count separately as addition and wrong output (Anderson); physicians self-evaluating their own clinic notes, a stated likely undercount (Taylor); and two adversarial model skeptics from different families with a tiebreak settling splits, which refuse roughly nine candidates in ten (this census). The spread above the rule, from 23.1\% to 86.3\%, therefore reflects the instruments at least as much as the scribes. Below the rule is the one comparison that shares a denominator, the share of notes carrying at least one omission: Taylor 18\% of 356 notes, ours 15.4\% [12.0, 19.0] of 565, counting a note as carrying an omission when at least one of its findings is placed under omission at tier 1. Different instruments, the same unit, and the two land together. Intervals are 95\% confidence intervals where the source study published one; ours, below the rule, resamples whole consultations. Anderson is the only share-of-errors figure with a published interval; the rest, ours included, are published without one.}
\label{fig:ledger}
\end{figure*}

Figure~\ref{fig:ledger} sets the published omission shares Section 4 reconciles beside the review standard that produced each, and Table~\ref{tab:comparators} lists the class mix, ours beside what each study reported. Our 23.1\% omission share sits below every published share of errors above, and the reason is the instrument, not a lower omission rate in these scribes. The panel verifies omission candidates at a quarter of the rate it verifies fabrications (6.1\% against 24.6\%); had omissions been verified at fabrication's rate, omission would be our largest verified class. Two counts of omission run through this paper and answer different questions. The first counts omissions after the open pass's survivors have been distributed into the published classes: 143 of the 618 verified findings, which is the count that matches published schemes and the one 23.1\% uses. The second is the omission hunt's own survival rate, its 78 verified findings out of the 1,284 candidates it proposed, which is the 6.1\%; Appendix A.9 keeps the two apart. The published studies used human reviewers applying their own thresholds to what should have been in the note - our panel was built to be stricter, though no shared items exist to measure the gap directly, and Section 4.1 shows how far a threshold alone can move a rate - and none of them applied the restatement criterion the previous subsection measured. Setting 23.1\% beside 54--86\% therefore compares instruments at least as much as scribes.

% T4 - comparator reconciliation. Generated by extract_p2_floats.py.
% Ours from master/findings_rates.json; published figures from docs/COMPARATOR-VERIFICATION.md.
\begin{table}[tp]
\centering
\small
\setlength{\tabcolsep}{5pt}
\begin{tabular}{@{}lccccc@{}}
\toprule
 & ours & Biro A & Biro B & Kernberg & Anderson\\
\midrule
Omission, share of errors & \textbf{23.1\%} & 83\% & 54\% & 86.3\% & 76.3\%\\
Addition & 29.3\% & 4\% & 11\% & 10.5\% & (in commission)\\
Wrong output & 33.5\% & 6\% & 10\% & 3.2\% & (in commission)\\
Irrelevant or misplaced & 7.4\% & 6\% & 25\% & not a class & not a class\\
Unmapped & 6.6\% & -- & -- & -- & --\\
\bottomrule
\end{tabular}
\caption{\textbf{Our verified error mix beside the published audits: the comparison is of instruments at least as much as of scribes.} Shares of errors per study; our shares are of 618 adversarially verified findings, the published shares are of human-reviewed error counts (Biro's two products, $n$=66 and $n$=61 errors; Kernberg, ChatGPT-4 prompted over encounter transcripts rather than a deployed scribe; Anderson, five platforms, omission 76.3\% [70.0, 83.3]). Our omission share sits far below every published share of errors, and not for want of hunting: omission is the largest of our eleven targeted discovery hunts (1,284 candidates, second only to the open pass), and it survives our panel at a quarter of fabrication's rate (6.1\% against 24.6\%, Table~\ref{tab:checkability}); at fabrication's survival rate it would be our largest verified class. Our denominator includes the 41 findings the placement matcher left unmapped, which the published shares have no counterpart for; excluding them would raise each of our four class shares by between half a point and two and a half points and change no ordering. On the one comparison sharing our denominator - share of \emph{notes} carrying the class - the two land together (15.4\% [12.0, 19.0] against 18\%), on different instruments. Anderson classifies errors as omission, commission or partially correct and publishes a share for omission alone; their commission covers our addition and wrong output (ours sum to 62.8\%), while their partially correct spans our omission and wrong output and cannot be split from either. Per-note rates are never set beside per-sentence, per-element or per-case rates. Two tier-2 categories - drug and code terminology, and biased or stigmatising language - were not hunted here, so they are not measured, which is not the same as zero.}
\label{tab:comparators}
\end{table}

The four-standard experiment says how much of such a spread the review layer alone can produce, because the class mix of each standard's verified findings is in its stored verdicts (Table~\ref{tab:standards}). On identical candidates the omission share of what gets verified runs from 22.3\% under the harsher skeptic to 33.9\% under the gentler, a 1.5-fold spread and 11.6 percentage points from the review layer alone, with the panel and the lenient instruction in between (Table~\ref{tab:standards}). Given both families the same strict instruction, that 11.6-point gap has a 95\% interval of {[}+2.5, +21.2{]}, which excludes zero; the pair that moves the instruction rather than the family does not separate on omission (Appendix A.3). The instruction's mark on the mix falls on a different class, and it is the largest single class move in the experiment: misplaced text runs at 6.7\% of the panel's verified findings against 18.0\% of the lenient reviewer's, on consultation-clustered intervals that do not overlap. The two levers therefore move different parts of the mix, and a published mix is sensitive to who reviewed as well as to how strictly.

The published audits' omission shares run from 54\% to 86\%. That is a 1.6-fold spread and 32 percentage points, the larger move in absolute terms because it sits at a much higher baseline than ours. It is also a spread across different corpora, products, reviewers and discovery processes as well as different standards, so it is not the same measurement as our four-standard contrast; what ours shows is that a spread of that order needs no product difference to arise. And the review layer is only half the instrument: the candidate pool these four standards reviewed - the sampled fifth of Section 4.1 - is 31.1\% omission once every candidate is placed at tier 1. A human reviewer reading for what is missing is a different discovery process, not just a different threshold: a published mix reflects what was hunted as well as what the review standard kept, and this experiment sizes only the second (Appendix A.3).

One published study counts the way we count. Taylor et al.~report the share of notes containing an accidental omission: 18\%. Ours, counted over the same unit - the share of notes carrying at least one verified finding placed under omission at tier 1, the 143-finding count of Appendix A.9 - is 15.4\% {[}12.0, 19.0{]}. Two counts built to be conservative land together, which is consistency rather than a validation of either: the instruments differ as much as any two in this paper, and only the unit is shared.

Two rules govern every cell of Table~\ref{tab:comparators}, and we commend them to anyone comparing scribe audits. First, denominators must match: our per-note rates are never set beside the per-sentence rates of CREOLA, the annotation platform of the sentence-level study in Section 2 \citep{asgari2025framework}, Anderson's 26.3\% of note elements, or Kernberg's 23.6 errors per case, because a note, a sentence, an element and a case are four different denominators. Second, a class that was not hunted is not measured, and is never zero: two tier-2 categories - drug and code terminology, and biased or stigmatising language - were left outside our discovery passes, and Table~\ref{tab:comparators}'s caption records them as not measured. Everything this census used to arrive at its numbers is in the release (Section 7), so a reader who thinks our standard too strict or too lenient can re-run it under their own, and any vendor's or auditor's rate can be asked to arrive the same way.

\section{What this means in practice}

Ambient scribes reach clinics today on the strength of the signing clinician's review, and the regulatory lines behind that differ by jurisdiction. In Great Britain, a scribe that summarises a consultation for clinician review is not a medical device, and the clinician's review is the mechanism that determination rests on \citep{mhra2026avt}. Australia's regulator draws the line at interpretation: a scribe that only transcribes clinical conversations sits outside the medical device framework, and one that generates a diagnosis or recommendation the clinician did not state sits inside it \citep{tga2026scribes}. In the United States, the FDA's regulatory approach to generative AI in medical devices is, as of August 2026, a discussion paper open for public comment \citep{fda2026genai}, while the Government Accountability Office finds few independent studies evaluating the accuracy of AI note-taking tools and reports that policymakers need more information about their performance to set the appropriate level of oversight \citep{gao2026notes}.

The concrete demand, where one exists, lands on the adopter: NHS England's guidance requires an adopting organisation to obtain the supplier's rate of errors, omissions and other dysfunctions, and to audit its clinical documentation on an ongoing basis \citep{nhsengland2026ambient}, and an MHRA sandbox workshop records a lack of established standards for benchmarking these products \citep{mhra2026airlockscope}. A rate is required in one jurisdiction and wanted in another, and none of the three supplies a standard for producing one; this census releases one.

\subsection{For the clinician who signs}

The verified failures are not spread evenly across the note. Four places hold most of the high-importance findings, and a fifth is the class this census most understates; those are the places to read closely before signing. \textbf{Allergy and medication state.} Does the note's list match what was actually said, at the stated dose, and does the note assert an allergy status the consultation never established? \textbf{Examination provenance.} Does the note claim an examination this consultation could not have contained? \textbf{Identity, where the record does not prefill it.} Is the patient's name and sex actually established anywhere? \textbf{The spoken diagnosis.} This was a single-product, format-level failure in this census, but it is cheap to check on any product. If you told the patient what you think this is, is that impression in the note? \textbf{Plans written as done.} Does the note record an investigation or a prescription as delivered when the consultation only proposed it? Two clusters of that shape hold 46 findings, 44 of them graded critical, from all three products - an ECG recorded as performed that was still to be booked, a choice between two antibiotics recorded as two issued prescriptions - and it is also the class our panel verifies least often (4.5\%, Table~\ref{tab:checkability}), so it is the count here with the most room beneath it. The failures easiest to miss are the fluent ones, and an invented calendar date reads as diligence rather than as an error. A reviewer scanning for what looks odd will not catch them - an inference from the form of these failures rather than something this census measured, since it counts what the products produced and not what a signing clinician catches (Section 6). Delegates at an MHRA workshop named the adjacent trap, a reliance bias in which clinicians check less as a product proves reliable \citep{mhra2026airlockscope}.

\subsection{For the buyer: three questions this census equips you to ask}

``What is your error rate?'' is the weak form of the question - this paper shows the answer can move by a factor of eight and a half at the candidate level, and from 28\% to 97\% of notes flagged, when the review standard changes and nothing else does. The strong forms, each answerable against our own record:

\textbf{Under what instrument, and may we see it?} A rate without its instrument cannot be compared with anyone else's, and the threshold can hide inside the headline: one vendor's published 90.2\% captured-entity rate counts only major or critical missing information as a defect, and nothing on the face of the number says so \citep{oleson2024deepscore}. Our own answer: every prompt, model version and verdict is public. The panel refutes roughly nine candidates in ten, and Section 4 shows what the same products score under a more lenient standard. Vendors have begun to concede the point in public: one now reports its error burden under two detection standards rather than one \citep{bergman2026quality}.

\textbf{Which of the failure clusters in this census has the vendor measured, and which fixed?} The seventeen subcategories in Figure~\ref{fig:ladder} and Table~\ref{tab:clusters} are a concrete checklist; four of them concentrate most of the high-importance findings, and two of those four carry findings from all three products we tested.

\textbf{What does this census find on your own consultations?} The pipeline is released and re-runnable with vendor accounts. On this corpus, discovery and verification cost about \$530 in model calls for 565 notes; the slow part is capture, because a listening product takes its audio in real time. Two things the cost does not show. Run on real consultations, the instrument sends each complete transcript and each complete note to two model providers, twelve discovery calls per note and two to three review calls per candidate, so an adopter needs the data-processing agreements and information-governance approval that implies before the first call; nothing in the pipeline needs to know who the patient is, and de-identified text serves every pass on one condition: the findings about invented names and sex compare what the note asserts against what the audio established, so transcript and note have to be de-identified together under one scheme that replaces names rather than blanking them. And size the run before commissioning it: a fifth of this census, 115 notes over 87 consultations, reproduced the note-level rate to within four points but with an interval from 20.5\% to 36.6\%, so a run of that size says roughly where a product sits and will not separate two products ten points apart.

\subsection{The gap a quality layer must fill, and only partly does}

The last practical question is whether the checking can be automated, since the volume ambient scribes enable is exactly what makes sign-off hard to sustain as a real check. The companion paper measures this directly, on a benchmark built from this study's corpus in which each omission test note has a named fact certainly missing. Its headline is an asymmetry: LLM judges - a second model that reads the transcript and the note and flags problems - separate notes with added or altered content from clean ones reliably (0.79 to 0.94 on a paired measure where 0.5 is a coin flip), and sit close to chance on content that is missing (0.50 to 0.63). At the reasoning budget judges normally run under - the cost and latency budget an evaluation actually runs to in the field, inside a deployed application - no rewording of the instructions repairs this.

Two methods then each recover part of the detection, and neither dominates the other. Both restructure the task the same way, listing the facts the transcript establishes and then checking each one against the note, and they differ in how. The first is the companion's enumerate-then-check pipeline, which runs the list and the checks as separate steps and flags the note if any critical fact is missing. That catches 24.6\% of notes containing an omission while wrongly flagging 2.7\% of clean notes, and every flag names the missing fact and how much it matters. The second is its evolved prompt, a single prompt written by an optimiser rather than by hand that carries out the same procedure inside one call when given room to reason, and it catches more - 36.9\% of notes containing an omission, and 58.9\% of the omissions the benchmark's severity rubric grades critical - at more noise (6.2\% of clean notes flagged) and roughly a tenth of the measured cost per note at one note per consultation (a third once the pipeline's per-consultation stages are amortised across several notes), with only a score to show for each flag. Both rates are measured on that benchmark, not on scribe output: on this census's own vendor notes neither method's threshold transfers, and each has to be re-established against a deployment's clean notes before its operating point means anything, though once re-established the evolved prompt still detects more than an ordinary judge at half its false-alarm rate. An omission whose content also appears somewhere else in the note is missed by both methods. So the error class the published audits find dominant is a class automated checking recovers only in part, and the failure map in Section 3 remains, for now, chiefly a human reviewer's job.

\section{Limitations}

\textbf{Not live traffic.} The consultations are recorded, simulated or authored, run through deployed products; no real patients and no live clinical workflow. The design gains its comparisons this way - identical consultations across three products, ground truth we control, no contamination of a production system - at the cost of ecological validity: real consultations are longer, messier and specialty-diverse in ways this corpus is not.

\textbf{No human comparator.} The census counts what three products wrote and has no clinician-written arm on the same consultations, so nothing here says whether a scribe note carries more failures than the note it replaces. A comparator would have to be counted by this same instrument to be readable beside these rates, since Section 4 shows what a different standard does to a count. The published two-arm comparisons make the point: a vendor evaluation flags 31\% of ambient notes against 20\% of physician-written ones for the same encounters, on a single binary rating item whose 20\% on human notes says as much about the instrument as about the notes \citep{palm2025assessing}, and a vendor preprint's paired simulation runs its own two detectors across both arms \citep{bergman2026quality}. Adoption turns on that contrast, and this study does not supply it.

\textbf{Capture paths differ by product.} Scribes B and C heard replayed audio; Scribe A consumed input through an API and contributed two templated notes per consultation. Between-product differences therefore carry a capture-path confound, which is one of the three reasons Section 3.2 gives for never ranking the products. Per-note rates are not normalised for note length either: a longer note carries more assertions and more chances to err, and the products' templates differ in length, though the one length contrast we can measure runs against it: Scribe A's detailed and short templates land at 0.447 and 0.461 findings per note (Table~\ref{tab:census}). What a product hears also bounds what it can omit: in a simulated medication-history study, video-enabled capture removed almost all omission errors relative to audio alone \citep{menz2026vision}.

\textbf{No patient-record context.} No product received demographics, a patient record, or an encounter date beyond what its own application supplies. Identity and date findings are therefore measured without the context some integrated deployments prefill; the behaviour that transfers is the products' filling of unknowns with invented specifics rather than flagging them. We put a size on it (Appendix A.7). The two classes are 137 of the 618 findings, 22.2\% of the census, and removing them while keeping every note in the denominator moves the headline from 31.3\% {[}27.0, 35.6{]} to 27.3\% {[}23.1, 31.5{]} without invented identity and 24.8\% {[}20.8, 29.0{]} without the dates as well. The two exclusions are kept apart on purpose and neither figure is the rate an integrated deployment would produce: a deployed scribe legitimately knows the encounter date and may resolve ``yesterday'' against it, where no product here had a record to draw a name or a sex from at all, and the exclusion assumes in any case that a record would have prevented every finding in both classes and nothing else. They are nested subsets of one sample, so neither figure may be tested against the other. The per-product effect is the part worth taking furthest: Scribe C drops 12.0 points on these exclusions and Scribe A 2.9, and which product reads worst inverts between the published rate and the rate net of these classes - neither gap separable at these intervals, and a fourth reason never to rank the products from this census.

\textbf{One clinical family.} UK primary care and US ambulatory encounters, in English. Nothing here measures specialty dictation, inpatient documentation, or other languages.

\textbf{Sign-off is not measured.} The census counts what the products produced, not what a signing clinician would have caught. That fluent failures survive review is an inference from their form, not a measurement here; a review-catch study on this corpus is the obvious next step.

\textbf{The instrument leans strict, and we measured by how much.} Section 4 puts a scale on the strictness that makes every rate here a floor for what the instrument missed: the same sampled candidates verify at 79.0\% under a lenient instruction against 10.3\% under the panel, and 96.5\% of sampled notes then carry at least one verified failure. The lenient reading has limits of its own - one call from one model family, no external ground truth to adjudicate between standards, and a candidate pool assembled by our own discovery passes - so the pair bounds the standard's contribution rather than supplying a second census. Discovery has the matching limit, and it is unmeasured: discovery is model-based, its sensitivity for errors no pass proposed is not estimated, and the discovering model's own importance filter dropped 7,780 of the 13,678 candidates before any skeptic saw them, a bin nothing in this paper audits. What discovery missed therefore makes every count here an undercount by an unknown amount. The judgement-call classes, omission above all, are held to a demanding bar, and the present-elsewhere share is published beside the omission rate as part of the instrument. The verification layer also upholds findings that sit early in the note more readily than late ones, and borderline so on transcript position - a position bias in verification, not discovery, detailed in Appendix A.5.

\textbf{One run of a stochastic instrument.} The panel ran once, at temperature 1.0 with fixed per-call seeds. Every interval here resamples consultations conditional on that single run; the instrument's own run-to-run variation is not measured, and a repeat run on a sample is the cheapest way to bound it. Until it is, a gap between two runs of the instrument should not be read as a product difference.

\textbf{Two classes not measured.} Drug and code terminology, and biased or stigmatising language, were deliberately not hunted; they are not measured and are not zero.

\textbf{A model-built census, human-checked at samples.} Discovery and verification are LLM passes; the human checks are blinded adjudications of sampled verified findings, of the panel's refusals and of the severity rubric's grading, plus unstructured spot-checks quantified nowhere (Appendix A.4). Most of that work is a physician author's, with the single-rater limits and the conflict that implies. The precision of the published set is measured at 95.2\% {[}77.3, 99.2{]} on the 21 sampled verified findings (20 stood), and on a second, disjoint sample of 12, adjudicated by the independent clinician, at 12 of 12, {[}75.8, 100{]}. The two are not pooled and neither is precise: 33 of the 618 findings have been formally adjudicated, both intervals are wide, and between them the two raters rejected one item in 46, so the sittings bound false positives without validating the refusals and without excluding simple acquiescence to a flag. Both sittings endorsed every panel-refused foil they saw, thirteen of thirteen. A calibration study with external reviewers over shared items (Section 7) is where a stronger estimate belongs. The countervailing protection is that every finding ships with its evidence quotes, so any reader can audit any finding against its transcript. One recent vendor preprint sharpens the human-check caveat: on the same AI notes, unaided clinician adjudication found error in 6.2\% where its calibrated automated reviewer found 24.4\% \citep{bergman2026quality}; if unaided review misses that much, a clinician's endorsement of a panel refusal is weaker evidence than it looks - and the same insensitivity is part of why sign-off needs the map this census provides.

\section{Conclusion and release}

One note in three from three deployed scribes carries a failure that survived a panel built to refute it, and the failures concentrate where the signing clinician is the only remaining check: allergy and medication state, identity, examination provenance, and - on one of the three products - the spoken diagnosis. Any such count is a joint property of the scribes and the instrument that produced it; we measured the instrument's share, and we publish both.

The release contains all 618 verified findings with transcript-side evidence quotes, panel verdicts, importance ratings and severity grades; the discovery and panel prompts; model versions and run manifests; the clustering recipe and its stability sweep; and the re-runnable pipeline, in the same repository and under the same DOI as the companion paper's benchmark (data at huggingface.co/datasets/ComposoAI/OmissionBench, code at github.com/composo-ai/omission-bench, DOI 10.5281/zenodo.22160954). The text of the notes the three products wrote is not released: the products' terms of service differ on republication, and we withhold all three alike rather than release asymmetrically, which would make products identifiable by their absence; released findings quote the transcript side, and a replicator with vendor accounts can regenerate equivalent notes with the released harness. The blinded sittings' concealed keys, answers and scored results are released; their item packs print the products' notes in full and are withheld with them. The products are anonymised as Scribes A, B and C throughout and we will not identify them.

These products update continuously, so this census is a snapshot they will outgrow. The instrument will not go out of date with them: run it on the next product generation, or on your own consultations, and the construction of the numbers will be the same, because the same published instrument will have produced them. How much two runs of that instrument differ on identical material is not measured here (Section 6). Because the products are not named and their notes are not released, the per-product figures in this paper cannot be independently re-derived; what is reproducible is the instrument, on any product a replicator holds an account with.

\textbf{What follows directly.} Four pieces of further work fall straight out of the census. A review-catch study on this corpus - which of these verified failures signing clinicians actually catch, at realistic review speed - would turn Section 5's inference about fluent failures into a measurement. A calibration study putting external human reviewers and the panel over shared items would measure directly the strictness gap Section 4 can currently only bound, and would give the agreement statistic the two disjoint sittings of Appendix A.4 cannot support. The two unhunted classes need terminology-grounded passes before they can be measured at all. And the census is built to be longitudinal: re-run on each product generation under the same released instrument, the rates become a trend rather than a snapshot.

\textbf{Author contributions.} S.F. designed the study, built and ran the census instrument, performed the analyses and the physician author's blinded sittings, and wrote the paper. L.M. reviewed the manuscript; R.L. and M.K. supported the work at Composo.

\textbf{Funding.} The study was carried out at Composo and received no external funding.

\textbf{Acknowledgements.} We thank Dr Hannah Warren-Miell for adjudicating the independent clinician's sitting of Section 3.1 blind; Dr Warren-Miell is not an author and had no other involvement in the study.

\textbf{Ethics and data.} The corpus contains no real patient data: PriMock57's consultations are between clinicians and actor patients (CC BY 4.0), ACI-Bench's encounters are simulated doctor-patient conversations released for research (CC BY 4.0), and the remaining strata are authored scenarios. No patients were involved and no clinical records were used.

\textbf{Competing interests.} The authors build evaluation tooling commercially. No commercial product of the authors is used anywhere in this study's measurement path; the census instrument is fully specified in the release. Section 5's advice concerns evaluation, which is the authors' commercial area; the instrument this paper recommends is released free, and following the advice requires no product of the authors.

\bibliography{references}

\clearpage
\appendix
\section{The census method in full}

This appendix sets out the material the main text compresses, in four parts. Part 1 is the taxonomy: every one of the 17 discovered subcategories with its size and composition, and the clustering recipe with its parameters and stability in full. Part 2 is the instrument: the behaviour of the verification panel, the controlled re-review of the census's candidates under four review standards, which kinds of claim survive verification, the instrument's own history, the audit of refused omission candidates, and the three blinded clinician sittings. Part 3 is the robustness checks: position bias, the trap-inflation check, and the census recomputed with the two classes a patient record would have prefilled removed. Part 4 is how to read the counts: the two published categories we did not hunt, and the counts that look alike but answer different questions, with the estimate of how many distinct errors the 618 findings represent.

Products are Scribe A, B and C throughout, the same anonymisation the main text uses. Scribe A is generated through an API at two note templates per consultation, so it contributes two notes per consultation where B and C contribute one; its columns below are counts of findings, not rates, and the per-note rates that account for the template difference are in Table~\ref{tab:census}.

\partdiv{Part 1 - The taxonomy}

\emph{What the 618 verified findings are, one cluster at a time, and how the clusters were made.}

\subsection{All 17 clusters}

The 618 verified findings were grouped by density on their descriptions, which put \textbf{563 of them into 17 clusters and left 55 unassigned}. Each cluster then sits under the fixed top two tiers of the scheme: the published four-class split at tier 1, and the mechanism level at tier 2. Sixteen clusters were placed under a tier-2 category; one, the retracted-device group, could not be placed under any category in the scheme and is reported in its own right.

Two importance columns appear below (salience as high / medium / low, severity as critical / supporting / peripheral) and they come from two different instruments, neither overwriting the other. \textbf{Salience} is the panel's own high, medium or low re-rating of each candidate it reviews - the released data's name for what the main text calls the panel's importance rating; survivors carry it into the released data. \textbf{Severity} is a written clinical rubric (released with the pipeline) applied to survivors only: critical, an error that would plausibly change clinical action or safety; supporting, one that degrades the record without changing what happens next; peripheral, no plausible consequence. A cluster can be high on one and lower on the other, and the two are not two readings of the same scale.

{
{\footnotesize\renewcommand{\arraystretch}{1.18}\begin{longtable}[]{@{}
  >{\raggedright\arraybackslash}p{(\linewidth - 20\tabcolsep) * \real{0.030}}
  >{\raggedright\arraybackslash}p{(\linewidth - 20\tabcolsep) * \real{0.260}}
  >{\raggedright\arraybackslash}p{(\linewidth - 20\tabcolsep) * \real{0.036}}
  >{\raggedright\arraybackslash}p{(\linewidth - 20\tabcolsep) * \real{0.082}}
  >{\raggedright\arraybackslash}p{(\linewidth - 20\tabcolsep) * \real{0.110}}
  >{\raggedright\arraybackslash}p{(\linewidth - 20\tabcolsep) * \real{0.028}}
  >{\raggedright\arraybackslash}p{(\linewidth - 20\tabcolsep) * \real{0.028}}
  >{\raggedright\arraybackslash}p{(\linewidth - 20\tabcolsep) * \real{0.028}}
  >{\raggedright\arraybackslash}p{(\linewidth - 20\tabcolsep) * \real{0.062}}
  >{\raggedright\arraybackslash}p{(\linewidth - 20\tabcolsep) * \real{0.108}}
  >{\raggedright\arraybackslash}p{(\linewidth - 20\tabcolsep) * \real{0.108}}@{}}
\toprule\noalign{}
\begin{minipage}[b]{\linewidth}\raggedright
\#
\end{minipage} & \begin{minipage}[b]{\linewidth}\raggedright
cluster
\end{minipage} & \begin{minipage}[b]{\linewidth}\raggedright
n
\end{minipage} & \begin{minipage}[b]{\linewidth}\raggedright
tier 1
\end{minipage} & \begin{minipage}[b]{\linewidth}\raggedright
tier 2
\end{minipage} & \begin{minipage}[b]{\linewidth}\raggedright
A
\end{minipage} & \begin{minipage}[b]{\linewidth}\raggedright
B
\end{minipage} & \begin{minipage}[b]{\linewidth}\raggedright
C
\end{minipage} & \begin{minipage}[b]{\linewidth}\raggedright
cons.
\end{minipage} & \begin{minipage}[b]{\linewidth}\raggedright
salience h / m / l
\end{minipage} & \begin{minipage}[b]{\linewidth}\raggedright
severity c / s / p
\end{minipage} \\
\midrule\noalign{}
\endhead
\midrule\noalign{}
\endfoot
\bottomrule\noalign{}
\endlastfoot
1 & Allergy status and medication list omissions & 111 & omission & omission & 19 & 40 & 52 & 33 & 71 / 26 / 14 & 96 / 15 / 0 \\
2 & Invented patient identity: name and sex & 93 & addition & fabrication & 10 & 45 & 38 & 34 & 49 / 15 / 29 & 79 / 14 / 0 \\
3 & Stated working diagnosis dropped from the note & 53 & omission & omission & 53 & 0 & 0 & 22 & 52 / 0 / 1 & 52 / 0 / 1 \\
4 & Relative timing converted to invented calendar dates & 44 & wrong output & timing & 0 & 4 & 40 & 17 & 2 / 16 / 26 & 12 / 32 / 0 \\
5 & Remote consult history written as objective examination & 42 & addition & examination provenance & 0 & 11 & 31 & 10 & 33 / 7 / 2 & 42 / 0 / 0 \\
6 & Planned investigations documented as already done & 32 & wrong output & modality hardening & 4 & 15 & 13 & 11 & 21 / 10 / 1 & 30 / 1 / 0 \\
7 & Fabricated or flipped symptom qualifiers (sleep, timing, location) & 28 & addition & fabrication & 0 & 14 & 14 & 12 & 11 / 13 / 4 & 19 / 9 / 0 \\
8 & Diabetes lab values distorted and re-provenanced & 21 & wrong output & dose value & 6 & 11 & 4 & 4 & 6 / 12 / 3 & 16 / 5 / 0 \\
9 & Fabricated calendar year for a vague onset date & 21 & wrong output & timing & 0 & 12 & 9 & 2 & 0 / 14 / 7 & 2 / 19 / 0 \\
10 & Phantom counselling and unasked negative history & 20 & addition & fabrication & 0 & 11 & 9 & 8 & 4 / 15 / 1 & 15 / 5 / 0 \\
11 & Family-history relative and lineage misassignment & 18 & wrong output & attribution & 0 & 0 & 18 & 1 & 4 / 11 / 3 & 11 / 7 / 0 \\
12 & Dropped work-absence and sick-note plan elements & 16 & omission & omission & 11 & 5 & 0 & 8 & 3 / 6 / 7 & 6 / 10 / 0 \\
13 & False denials of swelling or redness & 16 & wrong output & negation & 0 & 6 & 10 & 4 & 7 / 7 / 2 & 13 / 3 / 0 \\
14 & Alternative options recorded as issued prescriptions or orders & 14 & wrong output & modality hardening & 6 & 0 & 8 & 4 & 13 / 1 / 0 & 14 / 0 / 0 \\
15 & Unmeasured or corrupted values entered as objective vitals & 13 & addition & examination provenance & 0 & 1 & 12 & 4 & 7 / 4 / 2 & 11 / 2 / 0 \\
16 & Retracted device captured as delivered care & 11 & unplaced & unplaced & 7 & 4 & 0 & 1 & 5 / 6 / 0 & 11 / 0 / 0 \\
17 & Red-flag safety-netting downgraded to routine follow-up & 10 & wrong output & negation & 6 & 1 & 3 & 4 & 7 / 3 / 0 & 10 / 0 / 0 \\
& \textbf{clustered total} & \textbf{563} & & & \textbf{122} & \textbf{180} & \textbf{261} & & \textbf{295 / 166 / 102} & \textbf{439 / 122 / 1} \\
\end{longtable}\addtocounter{table}{-1}}
}

One member of cluster 6 has no rubric grade, so that row's severity counts to 31 of its 32 and the clustered severity total to 562 of 563. With the 55 unassigned findings added back (37 critical and 18 supporting, below), the 617 graded findings total 476 critical, 140 supporting and 1 peripheral.

The tier-2 names are plain-language forms of the released scheme's category identifiers, and two of them read differently from the table of hunted categories (Table~\ref{tab:checkability}): \emph{modality hardening} is the category that table lists as plans or possibilities written as done or definite, and \emph{examination provenance} is an examination finding recorded without a performed examination behind it.

The severity column's clustered total - 439 critical, 122 supporting, 1 peripheral - is a distribution to read against how the graded set was assembled, not as the severity mix of scribe errors at large. Findings reach the rubric only after two selections that each remove low-consequence material: the discovery model's own importance filter, which kept 5,898 of 13,678 candidates, and the panel, whose commonest refusal in the audited sample is that a candidate is not material (42.5\%) and whose survivors skew towards its own high importance rating (survival 15.7\% at high salience against 7.2\% at low). The survivors are therefore close to the population a clinical-consequence rubric is built to grade high, and the near-empty peripheral row says the pipeline discards peripheral material before grading reaches it, not that the scale discriminates three ways on this set. Two further cautions attach. The first is that the rubric is reproducible to within a grade but has no established direction: three blinded gradings now exist and they do not agree on which way the rubric leans. On the companion study's facts the machine graders sat about a grade more severe than the clinician; on sampled census survivors the author graded four of twenty above the rubric and none below; and on a fresh sample an independent clinician graded three of twelve below the rubric and none above (all three in A.4). Across the two census sittings 25 of 32 grades are exact and no disagreement anywhere exceeds one grade, so the disagreement that remains is between the raters rather than between rater and rubric, and \emph{no severity claim in this paper asserts a direction}. The second is that the grades count findings rather than distinct errors (A.9). What the distribution supports is that most verified findings are graded clinically consequential under the released rubric; it does not support a three-level severity discrimination among survivors, or any claim about the severity mix of scribe output before the filters.

The \textbf{55 unassigned findings} are not a residue of unimportant material: 22 are high salience, 27 medium and 6 low; 37 are graded critical and 18 supporting, and they span 20 consultations and all three products (A 6, B 35, C 14). They are findings whose descriptions sat too far from any dense group for the algorithm to place, and they are reported here rather than folded into a neighbouring cluster.

\emph{Read the consultation column before the size column.} Four clusters rest on very few consultations: family-history misassignment (18 findings, 1 consultation), the retracted device (11 findings, 1 consultation), the fabricated calendar year (21 findings, 2 consultations) and, less severely, the diabetes lab values (21 findings, 4 consultations). Repeated findings inside one consultation across products and templates inflate a count without adding independent evidence, so these are named as failure modes and never quoted as rates. The clusters that carry a rate are the seven spread across ten or more consultations, led by invented identity (34), allergy and medication (33), dropped working diagnosis (22) and invented dates (17).

\subsubsection{One example from each cluster}

Examples follow the main text's convention: note-side material is paraphrased from our verified finding records except short verbatim fragments in quotation marks; attribution gives the product and the source corpus.

\begin{enumerate}
\def\labelenumi{\arabic{enumi}.}
\tightlist
\item
  \textbf{Allergy status and medication list omissions.} Allergy status was asked and answered in the consultation and appears nowhere in the note, in an encounter where a drug is being prescribed. (Scribe B; PriMock57)
\item
  \textbf{Invented patient identity.} The patient's first name is inaudible in the recording and the note supplies a specific name and uses it throughout. (Scribe B; PriMock57)
\item
  \textbf{Stated working diagnosis dropped.} The clinician tells the patient the working diagnosis is gastroenteritis; the note contains no impression at all. (Scribe A, detailed template; PriMock57)
\item
  \textbf{Relative timing converted to invented dates.} The transcript gives only ``yesterday''; the note prints a specific calendar date. (Scribe C; PriMock57)
\item
  \textbf{Remote history written as objective examination.} A telephone consultation in which no examination took place, with examination findings documented in the note as observed. (Scribe C; PriMock57)
\item
  \textbf{Planned investigations documented as done.} The transcript has an ECG being arranged with reception if a slot is free and no result obtained during the consultation; the note records the ECG as performed. (Scribe C; authored)
\item
  \textbf{Fabricated or flipped symptom qualifiers.} The note asserts sleep disturbance the consultation never mentions. (Scribe B; PriMock57)
\item
  \textbf{Diabetes lab values distorted.} The patient recalls an HbA1c of about 7.7\%; the note records a definite 7.0\%. (Scribe B; ACI-Bench)
\item
  \textbf{Fabricated calendar year.} ``Last December'' becomes December 2025, a year the transcript never gives and one that sits oddly against the patient's own stated age. (Scribe C; ACI-Bench)
\item
  \textbf{Phantom counselling.} The transcript contains only a question about whether a previous clinician had discussed aspirin risks; the note records that the counselling was delivered in this encounter. (Scribe C; PriMock57)
\item
  \textbf{Family-history misassignment.} Breast cancer in the patient's maternal grandmother is recorded as the patient's mother's, contradicting the note's own history section. (Scribe C; ACI-Bench)
\item
  \textbf{Dropped work-absence plan element.} The clinician advises staying off work, which matters because the patient is a nurse, and the plan omits it. (Scribe A, detailed template; PriMock57)
\item
  \textbf{False denial of swelling or redness.} The patient reports inflamed knees and the examination records oedema and erythema; the note documents a denial of swelling and redness. (Scribe C; ACI-Bench)
\item
  \textbf{Alternative options recorded as issued.} A discussed choice between two antibiotics is recorded under prescriptions as though both were issued. (Scribe C; PriMock57)
\item
  \textbf{Unmeasured value entered as an objective vital.} A temperature the patient took at home enters the vitals section as an observed measurement; in this note the value itself is also corrupted, to an implausible 13.7°C against the spoken 37°C. (Scribe C; PriMock57)
\item
  \textbf{Retracted device captured as delivered care.} The clinician proposes a thumb spica, retracts it in the next breath and substitutes a wrist brace, and the note records the thumb spica as applied. The main text prints the verbatim note line and transcript line for this case. (Scribe A, detailed template; ACI-Bench)
\item
  \textbf{Red-flag safety-netting downgraded.} Sudden weight gain is given as a red flag warranting urgent contact, and the note reduces it to booking a routine appointment. (Scribe A, short template; authored)
\end{enumerate}

\subsection{The clustering recipe, in full}

Only the third tier of the scheme is discovered from the data. The recipe below produced it, and every parameter that could have been chosen otherwise is stated so the count can be reproduced or contested.

\begin{itemize}
\tightlist
\item
  \textbf{Text embedded}: the description field of each verified finding, prefixed with the fixed string \texttt{Failure\ mode\ described\ in\ this\ analysis:\textbackslash{}n} so that the embedding is oriented to the failure rather than to the clinical topic. The prefix is applied identically to every finding.
\item
  \textbf{Embedding model}: Cohere \texttt{embed-v4.0}. Embeddings are cached by content hash, so re-running the sweep or the labelling does not re-embed.
\item
  \textbf{Projection}: UMAP to 15 dimensions, metric cosine, \texttt{n\_neighbors} 15, random state 42, \texttt{n\_jobs} 1, \texttt{min\_dist} left at the library default.
\item
  \textbf{Clustering}: HDBSCAN with \texttt{min\_cluster\_size} 10 and \texttt{min\_samples} 3, excess-of-mass cluster selection, euclidean metric on the projected space, selection epsilon 0. Points HDBSCAN calls noise stay unassigned; nothing is forced into a nearest cluster.
\item
  \textbf{Labelling}: for each cluster, the members are ordered by distance from the L2-normalised medoid and the 20 nearest are shown to one call on the pinned labelling model at medium reasoning effort (pins in the released run manifests), which returns the cluster's name, its description, and a proposed placement in the scheme. One call per cluster, 17 calls in total.
\item
  \textbf{Placing a cluster in the scheme}: a cluster goes under the tier-2 category that at least 60\% of its members already carry from the pass that found them. Where the members do not agree that strongly, the labelling call places it; where the call declines, the cluster is reported as unplaced, in its own row of the table above and its own subsection of the main text (Section 3.5). In this run \textbf{no cluster reached the 60\% member threshold} - the highest member agreement is 53\%, in the dropped-working-diagnosis cluster, and the median is 30\% - so the labelling call placed all 16 placed clusters and declined on the retracted-device cluster. The member vote decided no placement, so every placement is the labelling call's; the clusters are groups of similar clinical situations, not re-derivations of the pass that found them.
\item
  \textbf{Inputs recorded}: scheme definition file version 1.0, with its sha256 in every run manifest, over 618 verified findings, yielding 17 clusters and 55 noise points.
\end{itemize}

\subsubsection{Seed stability, and what the parameter grid looked like}

Re-running the projection at UMAP seeds 42, 43 and 44 with every other parameter held gives \textbf{17, 20 and 21 clusters}, with pairwise adjusted Rand index \textbf{0.667, 0.707 and 0.808, mean 0.727}, where 1.0 is identical groupings and 0 is the agreement two random partitions would reach by chance. The count is therefore stable only to within a few, and the boundaries are largely but not exactly reproducible. The main text states the count with that instability attached, and any use of these clusters should keep it attached.

We also swept the three parameters that most change the answer, over 24 configurations: \texttt{n\_neighbors} in \{10, 15\}, \texttt{min\_cluster\_size} in \{10, 8, 6, 5\} and \texttt{min\_samples} in \{1, 2, 3\}. Cluster counts across the sweep run from \textbf{17 to 54} and unassigned findings from \textbf{33 to 119}. The sweep script, written before any cluster's contents were read, flagged a configuration as good if it gave between 6 and 16 clusters with no cluster over 45\% of the pool, and no configuration in the grid met that band. We kept the one that came closest, which is also the coarsest corner of the grid: 17 clusters with 55 findings unassigned. Every finer setting fragments the same material rather than reorganising it, and the 111-member allergy and medication group survives intact in that configuration alone, splitting into smaller pieces in all 23 others. The largest cluster in the kept configuration therefore holds 111 of the 618 findings, 18.0\%, so no single group dominates the taxonomy.

\partdiv{Part 2 - The instrument}

\emph{What the verification panel does to a count: its behaviour, the same candidates under four review standards, which kinds of claim survive it, its own history, the audit of its refusals, and the three blinded clinician sittings.}

\subsection{What the verification instrument contributes to a failure rate}

\subsubsection{The panel, and how it behaves}

Every candidate finding that cleared the discovery-stage importance filter - the discovering model's own high and medium ratings, 5,898 of the 13,678 candidates it produced - went to two skeptics drawn from different model families. Each was shown the full note and the full transcript and told to refute the finding if refuting it was at all defensible. Both keeping means verified; both refuting means cut; a split goes to one tiebreak call whose verdict is final. An unparseable or failed reply counts as a refutation. Every candidate the panel reviews also receives the panel's own salience re-rating, whatever its verdict; survivors additionally get a severity grade against the written rubric.

The models, named (exact pins, provider routing and settings are in the released run manifests; the harsher skeptic's calls ran at temperature 1.0 with medium reasoning effort, a 6,000-token cap and deterministic per-call seeds). Discovery, its importance filter, the salience re-rating, the severity grades and the cluster labelling call (A.2) all run on \texttt{anthropic/claude-opus-5}; the embeddings under the clustering are Cohere \texttt{embed-v4.0}. The two skeptics are \texttt{anthropic/claude-opus-5} (the harsher) and \texttt{openai/gpt-5.5} (the gentler); the tiebreak is \texttt{openai/gpt-5.4} at high reasoning effort; the refusal audit's classifier (A.4) is \texttt{openai/gpt-5.5}; the lenient re-review below runs on \texttt{anthropic/claude-opus-5}, the same pin as the harsher skeptic. The harsher skeptic is the same model that generated the candidates, so half the panel reviews its own proposals - and it is the harsher half, refuting 90.6\% of what it sees against the other family's 81.3\% on the same pool, so self-leniency is not what the stored votes show.

The two skeptics do not refute at the same rate: \textbf{the harsher refutes 90.6\% of what it sees and the gentler 81.3\%}, over 5,898 candidates each. That asymmetry is why a single dissent cannot fail a finding on its own. Under a three-member panel a failed or unparseable reply is a silent cut; under two members plus a tiebreak it produces a split, which gives a real third opinion on the evidence. The panel's decisions divide as \textbf{4,663 unanimous refutals, 812 splits sent to the tiebreak, and 423 unanimous keeps} (4,663 + 812 + 423 = 5,898), and of the 812 tiebreaks \textbf{195 upheld the finding and 617 cut it}, so the 618 verified findings are 423 unanimous keeps plus 195 tiebreak upholds. Unaided agreement between the two families is \textbf{86.2\%}.

The tiebreak model was drawn from the same developer as the gentler skeptic, so the run records which way it went on every split: it \textbf{sided with the harsher skeptic 531 times and the gentler 281}, so it did not back its own family.

The panel also re-rates importance, and it moves a lot. Discovery had called the 5,898 candidates 1,211 high and 4,687 medium; the panel called them \textbf{2,014 high, 2,377 medium and 1,507 low}. Every salience figure in this paper is the panel's. Survival rises with the panel's own rating: \textbf{high 317/2,014 = 15.7\%, medium 193/2,377 = 8.1\%, low 108/1,507 = 7.2\%}, so the panel is roughly twice as likely to uphold a finding it rates high.

What is published alongside the rate: the discovery prompt templates are carried over verbatim from the earlier version of the instrument, with nine of the eleven targeted pass instructions pulled from that module at import so the scheme cannot quietly reword them; the panel prompt is the same sentences reordered so the transcript comes first, with an assertion at start-up that fails the run if it is ever anything but a reordering; the scheme's definition file carries a version and a hash in every run manifest; and both skeptics see an identical prompt, so the cross-family comparison is not confounded by prompt shape.

\subsubsection{The same candidates under four review standards}

The controlled version of the instrument question holds everything fixed except the review standard, on a stratified sample of the census's own candidates (Section 4.1, Table~\ref{tab:standards}). This subsection sets out its method and full read-out; the instrument's own history, whose rows differ in model generation, family mix, discovery depth and corpus at once, closes A.3.

\textbf{The sample.} 115 of the 565 notes, drawn as whole notes with strata of product by source, proportional with largest-remainder allocation, from a fixed seed. The draw yields 1,295 of the census's 5,898 candidates (22.0\%; the note-level sampling fraction is 20.3\%) over 114 notes and 87 consultations, with one drawn note carrying no candidate and remaining in the note-level denominator. Whole notes rather than candidates were drawn so that the share of notes with at least one verified finding could be estimated without bias. The cost on the candidate-level tables is small - the worst marginal gap against the census across category, salience, product and source is 1.84 percentage points - and each standard reproduces its census-wide rate on the sample: the panel 10.35\% against 10.48\%, the harsher skeptic alone 9.34\% against 9.41\%, the gentler 17.76\% against 18.70\%. The draw, its identifiers and their hashes are in the release and reproduce from the seed alone. One category, laterality, has only five sampled candidates and is excluded from per-category read-outs.

\textbf{The four standards.} The census panel's stored votes contain each skeptic's solo verdict on every candidate before any tiebreak, so three standards are re-read from the census run itself: the harsher-family skeptic alone, the gentler-family skeptic alone, and the full panel (both keep, or a split goes to the tiebreak). The fourth was newly run: one call per candidate on the same model, transport and settings as the panel's harsher skeptic, with the evidence sections of the prompt identical by construction - a runtime assertion fails the run if they differ - and only the review instruction and its closing question replaced. The strict instruction reads:

\begin{quote}
Your job is to REFUTE it if at all defensible - be a hard skeptic. It is NOT a real, reportable discrepancy if it is any of: paraphrase, clinical synonymy, a standard abbreviation, a normal documentation convention, justified clinical inference, a quote that does not actually appear, a mischaracterisation by the tool, or clinically immaterial. Default to REFUTED if you are uncertain. {[}\ldots{]} Is this a REAL, correctly-characterised discrepancy of the stated kind?
\end{quote}

The lenient instruction reads:

\begin{quote}
Your job is to apply the standard an informal reviewer would use before signing the note off: would you flag this back to the person who wrote it? It IS worth reporting if a reasonable reviewer would want it corrected before the note enters the record, even if it is minor, arguable, or a matter of documentation practice rather than clinical fact. {[}\ldots{]} Is this a genuine documentation error worth reporting to the note's author?
\end{quote}

Both prompts are released verbatim, with their hashes recorded in every run manifest. The strict list's exclusions describe defects of the candidate, not of the note: a candidate whose claimed quote does not actually appear, or that mischaracterises what it flagged, fails on its own evidence. Of the 1,295 lenient replies, 1,294 parsed cleanly and one was recovered by a regex over the model's own true-or-false token, marked as salvaged in the released record; four calls whose replies initially failed to parse were re-run at their fixed seeds, and the run closed with no transport errors and no unreadable verdicts. The stored strict verdicts carry the census run's failed-reply-counts-as-refute rule; that run recorded a single defaulted vote across its 11,796 skeptic replies, so parse failure contributes at most one refusal to the strict rows.

\textbf{Candidate-level read-out}, with Wilson intervals first and consultation-clustered bootstrap intervals second (87 clusters, 10,000 draws):

{
{\footnotesize\renewcommand{\arraystretch}{1.18}\begin{longtable}[]{@{}
  >{\raggedright\arraybackslash}p{(\linewidth - 8\tabcolsep) * \real{0.2000}}
  >{\raggedright\arraybackslash}p{(\linewidth - 8\tabcolsep) * \real{0.2000}}
  >{\raggedright\arraybackslash}p{(\linewidth - 8\tabcolsep) * \real{0.2000}}
  >{\raggedright\arraybackslash}p{(\linewidth - 8\tabcolsep) * \real{0.2000}}
  >{\raggedright\arraybackslash}p{(\linewidth - 8\tabcolsep) * \real{0.2000}}@{}}
\toprule\noalign{}
\begin{minipage}[b]{\linewidth}\raggedright
standard
\end{minipage} & \begin{minipage}[b]{\linewidth}\raggedright
verified
\end{minipage} & \begin{minipage}[b]{\linewidth}\raggedright
share
\end{minipage} & \begin{minipage}[b]{\linewidth}\raggedright
Wilson 95\%
\end{minipage} & \begin{minipage}[b]{\linewidth}\raggedright
clustered 95\%
\end{minipage} \\
\midrule\noalign{}
\endhead
\midrule\noalign{}
\endfoot
\bottomrule\noalign{}
\endlastfoot
harsher skeptic alone, strict & 121/1,295 & 9.34\% & {[}7.88, 11.05{]} & {[}5.97, 12.90{]} \\
gentler skeptic alone, strict & 230/1,295 & 17.76\% & {[}15.78, 19.94{]} & {[}13.14, 22.51{]} \\
full panel with tiebreak & 134/1,295 & 10.35\% & {[}8.80, 12.13{]} & {[}6.36, 14.65{]} \\
same model as the harsher skeptic, lenient & 1,023/1,295 & 79.00\% & {[}76.69, 81.13{]} & {[}75.82, 81.97{]} \\
\end{longtable}\addtocounter{table}{-1}}
}

The controlled contrast, lenient against strict on the same model: ratio 8.45 (clustered 95\% {[}6.15, 13.09{]}), difference +69.7 percentage points {[}+65.5, +73.6{]}; exact McNemar over the paired candidates, 902 discordant one way and 0 the other, p = 5.9e-272. Lenient against the full panel: 7.63x {[}5.42, 12.36{]}. Panel against the harsher skeptic alone - the machinery at a fixed instruction: +1.0 percentage point, clustered {[}-1.18, +3.67{]}, 38 and 25 discordant, exact McNemar p = 0.13. Gentler against harsher at the fixed strict instruction: +8.4 points, clustered {[}+5.0, +12.1{]}. Of the 1,295 candidates, both standards keep 134, the lenient standard alone keeps 889, the panel alone keeps none, and both cut 272: the lenient standard keeps 100\% of the panel's survivors, which is the nesting result in the main text.

\textbf{Note-level read-out} - the bounded census statement - over all 115 drawn notes:

{
{\footnotesize\renewcommand{\arraystretch}{1.18}\begin{longtable}[]{@{}
  >{\raggedright\arraybackslash}p{(\linewidth - 8\tabcolsep) * \real{0.2000}}
  >{\raggedright\arraybackslash}p{(\linewidth - 8\tabcolsep) * \real{0.2000}}
  >{\raggedright\arraybackslash}p{(\linewidth - 8\tabcolsep) * \real{0.2000}}
  >{\raggedright\arraybackslash}p{(\linewidth - 8\tabcolsep) * \real{0.2000}}
  >{\raggedright\arraybackslash}p{(\linewidth - 8\tabcolsep) * \real{0.2000}}@{}}
\toprule\noalign{}
\begin{minipage}[b]{\linewidth}\raggedright
standard
\end{minipage} & \begin{minipage}[b]{\linewidth}\raggedright
notes with at least one verified finding
\end{minipage} & \begin{minipage}[b]{\linewidth}\raggedright
share
\end{minipage} & \begin{minipage}[b]{\linewidth}\raggedright
Wilson 95\%
\end{minipage} & \begin{minipage}[b]{\linewidth}\raggedright
verified findings per note
\end{minipage} \\
\midrule\noalign{}
\endhead
\midrule\noalign{}
\endfoot
\bottomrule\noalign{}
\endlastfoot
harsher skeptic alone, strict & 32/115 & 27.8\% & {[}20.5, 36.6{]} & 1.05 \\
gentler skeptic alone, strict & 63/115 & 54.8\% & {[}45.7, 63.6{]} & 2.00 \\
full panel with tiebreak & 32/115 & 27.8\% & {[}20.5, 36.6{]} & 1.17 \\
same model, lenient & 111/115 & 96.5\% & {[}91.4, 98.6{]} & 8.90 \\
\end{longtable}\addtocounter{table}{-1}}
}

\textbf{Per-category and per-product behaviour under the lenient standard}, which Section 4.2 draws on: rank correlation with the panel's category ordering 0.72 over the eleven categories with at least twenty sampled candidates, the category spread compressing from eleven-fold to one-and-a-half-fold, misplaced text moving from 4.3\% to 88.5\%, the salience spread compressing from 3.2-fold to 1.2-fold, and the between-product spread from 4.5-fold (Scribe C highest) to 1.16-fold (Scribe B highest, Scribe A lowest under both standards).

\textbf{The class mix of each standard's survivors} (Section 4.4; every survivor placed at tier 1 by the same deterministic matcher as A.9, one code path). Omission is 22.3\% of the harsher strict skeptic's verified findings, 33.9\% of the gentler's, 23.9\% of the panel's and 28.2\% of the lenient re-review's - a 1.52-fold spread on identical candidates. Misplaced or irrelevant text moves most: 6.7\% of the panel's survivors against 18.0\% of the lenient re-review's, consultation-clustered 95\% intervals {[}2.1, 11.5{]} and {[}16.0, 20.2{]}. Given both families the same strict instruction, the gentler family's verified findings carry an omission share 11.6 percentage points above the harsher family's, 95\% interval {[}+2.5, +21.2{]}, which excludes zero. The panel-against-lenient omission pair does not separate: 23.9\% {[}13.0, 39.6{]} against 28.2\% {[}24.0, 33.1{]}, overlapping intervals. The candidate pool the standards review is itself 31.1\% omission at tier 1, and a standard that verifies most of the pool - the lenient re-review keeps 79\% - is pulled towards the pool's mix by construction, so strict standards are where the mix can move furthest.

\textbf{Scope.} The three strict standards are re-readings of stored verdicts, and the lenient re-review is one call per candidate from one model family. No external ground truth adjudicates between standards - the physician review in A.4 calibrates the strict panel's refusals only. And the candidate pool is the one our discovery passes assembled, so all four rates are shares of that pool, not of all possible failures. The record store, the draw with its hashes, the analysis file and the per-run manifests with both prompts verbatim are in the release.

\subsubsection{What survives is what can be checked}

The main text's Table~\ref{tab:checkability} lists survival by tier-2 category. By the top tier the same figures are: addition 115/620 = 18.5\%, open pass 196/1,396 = 14.0\%, wrong output 183/1,711 = 10.7\%, omission 78/1,284 = 6.1\%, and irrelevant or misplaced text 46/887 = 5.2\%.

\emph{Survival tracks how objectively checkable a claim is rather than whether it is an omission.} Section 4.2 sets that ordering out over the tier-2 categories, with each category's rate in Table~\ref{tab:checkability}; the tier-1 figures above are the same behaviour one tier up. \emph{Omission is not the lowest-surviving category}, which matters because that is the easy misreading of the table.

\subsubsection{The instrument's own history: the same corpus under four panels}

The controlled experiment above holds everything fixed but the review standard. The instrument's own history points the same way with far less control, and is set out here as corroboration: four panel designs ran on the same task shape over the same three products, and what moved between each pair of rows is stated.

{
{\footnotesize\renewcommand{\arraystretch}{1.18}\begin{longtable}[]{@{}
  >{\raggedright\arraybackslash}p{(\linewidth - 10\tabcolsep) * \real{0.1667}}
  >{\raggedright\arraybackslash}p{(\linewidth - 10\tabcolsep) * \real{0.1667}}
  >{\raggedright\arraybackslash}p{(\linewidth - 10\tabcolsep) * \real{0.1667}}
  >{\raggedright\arraybackslash}p{(\linewidth - 10\tabcolsep) * \real{0.1667}}
  >{\raggedright\arraybackslash}p{(\linewidth - 10\tabcolsep) * \real{0.1667}}
  >{\raggedright\arraybackslash}p{(\linewidth - 10\tabcolsep) * \real{0.1667}}@{}}
\toprule\noalign{}
\begin{minipage}[b]{\linewidth}\raggedright
panel design
\end{minipage} & \begin{minipage}[b]{\linewidth}\raggedright
discovery
\end{minipage} & \begin{minipage}[b]{\linewidth}\raggedright
corpus
\end{minipage} & \begin{minipage}[b]{\linewidth}\raggedright
candidates
\end{minipage} & \begin{minipage}[b]{\linewidth}\raggedright
verified
\end{minipage} & \begin{minipage}[b]{\linewidth}\raggedright
survival
\end{minipage} \\
\midrule\noalign{}
\endhead
\midrule\noalign{}
\endfoot
\bottomrule\noalign{}
\endlastfoot
three skeptics, one model family, earlier model generation & 10 passes per note & 121 notes, pilot corpus & 466 & 174 & \textbf{37.3\%} \\
three skeptics, one model family, current generation & 10 passes per note & 6 notes & 48 & 3 & \textbf{6.25\%} \\
two skeptics, different families, plus tiebreak & 12 passes per note & 9 notes & 66 & 14 & \textbf{21.2\%} \\
the final panel: two skeptics, different families, plus tiebreak & 12 passes per note & 565 notes, 142 consultations & 5,898 & 618 & \textbf{10.48\% {[}8.9, 12.1{]}} \\
\end{longtable}\addtocounter{table}{-1}}
}

Only the last step in this table isolates a single change. Between the first two rows the model generation and the corpus move together, from 121 pilot notes to 6, and survival falls from \textbf{37.3\% to 6.25\%}: the earlier panel refuted 62.7\% of what it saw, the current-generation panel of the same shape 93.8\%. The next step moves three things at once - the panel architecture, the discovery instrument from ten passes to twelve, and the corpus from 6 notes to 9 - and takes survival back up to 21.2\% on 66 candidates. Only the last step holds the instrument fixed and moves one thing, from 9 notes to all 565, which takes survival to \textbf{10.48\%}, a factor of three and a half below the 37.3\% the earliest instrument would have reported on the same kind of work. The scribes did not change across any of this.

The two middle rows rest on small samples: 3 of 48 and 14 of 66 are counts a single consultation's worth of material can move, and they are reported as the instrument's history rather than as measurements of anything about the products. The final panel's row is the one with an interval on it. One further basis note: the pilot row's candidate pool predates the importance filter the final run applies; restricted to the candidates the discovering model rated high or medium - the final filter's rule - the pilot panel's survival is 148 of 361 = 41.0\%, so the filtered-basis contrast is slightly larger than the headline pair.

\subsection{The refusal audit, and the blinded clinician sittings}

A near-zero omission rate could mean the scribes rarely omit important material or that the instrument suppresses omissions, and the rate alone cannot say which; so we audited the refusals. A separate model pass - the auditor role, \texttt{openai/gpt-5.5}, never the model that generated the candidate, though it is the gentler skeptic in a second job, so on every unanimous refusal it re-read a verdict it had itself cast - re-read a stratified sample of \textbf{40 refused omission candidates}, this time with the full note in the prompt, and classified each refusal.

{
{\footnotesize\renewcommand{\arraystretch}{1.18}\begin{longtable}[]{@{}
  >{\raggedright\arraybackslash}p{(\linewidth - 8\tabcolsep) * \real{0.2000}}
  >{\raggedright\arraybackslash}p{(\linewidth - 8\tabcolsep) * \real{0.2000}}
  >{\raggedright\arraybackslash}p{(\linewidth - 8\tabcolsep) * \real{0.2000}}
  >{\raggedright\arraybackslash}p{(\linewidth - 8\tabcolsep) * \real{0.2000}}
  >{\raggedright\arraybackslash}p{(\linewidth - 8\tabcolsep) * \real{0.2000}}@{}}
\toprule\noalign{}
\begin{minipage}[b]{\linewidth}\raggedright
bucket
\end{minipage} & \begin{minipage}[b]{\linewidth}\raggedright
released label
\end{minipage} & \begin{minipage}[b]{\linewidth}\raggedright
n
\end{minipage} & \begin{minipage}[b]{\linewidth}\raggedright
share
\end{minipage} & \begin{minipage}[b]{\linewidth}\raggedright
is the refusal clean?
\end{minipage} \\
\midrule\noalign{}
\endhead
\midrule\noalign{}
\endfoot
\bottomrule\noalign{}
\endlastfoot
Not material: a salience call the panel is entitled to make & \texttt{not\_material} & 17 & 42.5\% & yes \\
The fact is stated somewhere else in the note & \texttt{present\_elsewhere} & 10 & \textbf{25.0\%} & \textbf{no} \\
Not required in a note of this kind: a scope call, not an omission & \texttt{not\_required} & 9 & 22.5\% & arguably \\
Wrongly cut: a real, reportable omission the panel refused & \texttt{wrongly\_cut} & 3 & 7.5\% & no \\
Spurious: the transcript never contained the fact & \texttt{spurious} & 1 & 2.5\% & yes \\
\textbf{total} & & \textbf{40} & \textbf{100\%} & \\
\end{longtable}\addtocounter{table}{-1}}
}

The audit's headline is the \textbf{25.0\% present-elsewhere share}: in a quarter of the sampled refusals the panel refused an omission candidate because the fact appears somewhere else in the note. Clinical notes are redundant documents, and the panel counts a restatement anywhere as capture - which is not necessarily the clinical sense the published taxonomies mean, since a fact stated outside the section a reader would consult is closer to what those taxonomies call misplaced text. This is a live caveat on our omission rate rather than a settled one. A pilot version of the same audit on 15 refusals had measured the share at 0\%; at scale it is a quarter, which is a reminder of how little a check of that size can rule out.

The other end of the table is the panel's own record on false kills. Only 3 of 40 refusals are candidates the classifier judged wrongly cut, so the panel's characteristic failure is counting a restatement as capture, not refusing real omissions. And omission survival by the panel's own salience rating is \textbf{high 52/340 = 15.3\%, medium 20/580 = 3.4\%, low 6/364 = 1.6\%}, over the 1,284 omission candidates that reached the panel. High-salience omissions survive at essentially the all-category high-salience rate of 15.7\%, so the low overall omission rate is the panel making importance judgements on medium and low material rather than a bar set to suppress the study's own headline.

\subsubsection{What the refusal-and-rubric sitting did and did not settle}

Ten of these 40 refusals went into a structured, blinded validation sitting by a physician author (they are an author; that conflict applies to every number here). The ten were stratified over the audit's own buckets - all three candidates the auditor had called wrongly cut, three from the present-elsewhere bucket, two not-material and two not-required - so the pack is enriched for false kills by design; they were interleaved so the bucket labels never leaked, with the question ``reading the note, is this a real omission the study should have counted?''. The same sitting also graded twenty severity facts, and this appendix calls it the refusal-and-rubric sitting throughout; the precision sitting below is a separate, later one.

\textbf{On false kills the sitting is supportive.} The physician author called none of the ten a reportable omission: 8 fair cuts, 2 abstentions and 0 real omissions. Counting the two abstentions in the denominator, that is 0 of 10. Because the ten were selected to include every case the model auditor doubted, no false-kill rate is estimated from them; what the sitting shows is that a clinician reading a pack drawn to over-represent the panel's likely mistakes found no reportable omission in it.

\textbf{On the classifier's own bucketing the sitting disagrees.} Agreement with the audit's buckets is \textbf{5 of the 8 answered items, 62.5\%, Wilson 95\% {[}30.6, 86.3{]}}, and all three disagreements are exactly the three \texttt{wrongly\_cut} items: an omitted instruction to book a follow-up appointment with the patient's own GP within a few days, an omitted metformin dose, and an omitted description of a planned operation and the reason for its urgency. The clinician called all three fair cuts. The audit's 7.5\% wrongly-cut share should therefore be read as an upper bound produced by a model, not as a measured false-kill rate; on the only sample a clinician has seen, the model auditing the panel was harsher on the panel than the doctor was.

\textbf{On the present-elsewhere caveat the sitting settles nothing, and that is the gap that matters.} Of the three items from that bucket in the pack, they answered one, agreeing that cut was fair, and abstained on the other two. The 25\% present-elsewhere share is therefore essentially unvalidated by the sitting, and the sitting must not be cited in support of it. Both abstentions are a property of the pack: eight of the ten items printed only the part of the note covering the point and two printed the note in full, and a present-elsewhere item cannot be judged without the whole note, which is what one of the two abstention comments says in as many words. The denominator for this section of the sitting is therefore 8, and the eight answers were given without the whole note in view, so on those items ``fair cut'' means the fact did not have to be in the note, not that it was verified as being elsewhere in it. The sitting's per-item answers and key, and the further stages it covered for the companion paper's instruments, are in the release; the item pack itself prints the products' notes and is withheld with them (Section 7).

\subsubsection{The severity rubric, calibrated blind in the same sitting}

Every verified finding carries a severity grade against the written clinical rubric, so the rubric's grading is the instrument's second judgement layer, and the same blinded sitting put the physician author under it: twenty facts graded blind against the rubric, drawn from the companion study's severity strata. The rubric is the same written instrument that grades this census's survivors, applied there as a consensus of two graders from different model families, with disagreements resolved to the lower grade, and here at verification time - so the sitting validates the grading axis, not the individual grades on these 618 findings.

\begin{itemize}
\tightlist
\item
  \textbf{Exact agreement 14/20 = 70.0\%, Wilson 95\% {[}48.1, 85.5{]}}; counting an adjacent grade as half, \textbf{85.0\%}; linear-weighted kappa \textbf{0.63} (bootstrap 95\% interval 0.32 to 0.86, indicative at n=20).
\item
  No disagreement exceeded one grade: six adjacent, none further.
\item
  The disagreement concentrates in the machine's critical row: of the seven facts the machine consensus graded critical they kept three, moving four down to supporting; the other two disagreements ran upward. On this sample the rubric's machine graders sat about one grade more severe than the clinician at the top of the scale. The precision sitting and the independent clinician's sitting below disagree with each other on direction, so this is one grading of three and establishes no direction; A.1 reads the three together.
\end{itemize}

The disclosures that apply to every number in this section: the reviewer is an author, they wrote the rubric they graded against, and there was one rater per sitting with no re-test. Beyond the structured sittings they also spot-checked verified findings across the wider set - unstructured review, reported as such and quantified nowhere.

\subsubsection{The precision sitting: sampled verified findings adjudicated blind}

The sections above audit what the panel refused; this one audits what it published. A refuted real error can only lower the count, but an upheld candidate the transcript does not support raises it, and nothing in the floor argument bounds that direction - so the precision of the verified set was measured directly.

\textbf{Design.} 30 items, adjudicated blind by the physician author: 21 verified findings drawn uniformly at random from the 618, interleaved with 9 foils drawn uniformly from the high-salience refused stratum - candidates the panel refused but rated high salience, chosen so the foils are not trivially distinguishable from survivors. Order shuffled from a fixed seed; no status label anywhere in the pack. The pack was drawn at 45 items - 30 verified findings and 15 foils - and adjudication stopped after the first 30 of that shuffle, which is itself a random draw; the prefix held the 21 verified findings over 19 consultations and the 9 foils above, and the sitting is reported on those 30. One limit of the foil half of the pack is stated up front: the refusal audit above finds only 2.5\% of refusals spurious, and the high-salience stratum is where genuine-but-refused material concentrates, so no item in the pack was expected to be baseless - the sitting tests whether a clinician endorses what the pipeline produced and what it refused, not whether the clinician rejects planted decoys, which the design does not contain. Each item presented the candidate's claimed failure with its evidence quotes, then the full note and the full transcript - the refusal-and-rubric sitting printed only the part of the note covering the point in eight of its ten items, which is why it could not settle the present-elsewhere bucket, and that limit is not repeated here. Two questions per item: is this a genuine documentation failure a reviewing clinician should accept (genuine / not genuine / cannot judge), and, on genuine verdicts only, a severity grade against the written rubric.

\textbf{Precision.} 20 of the 21 verified findings stood: \textbf{95.2\%, Wilson 95\% {[}77.3, 99.2{]}} (the 21 findings span 19 consultations; no abstentions; a consultation-clustered interval is in the released results but degenerates at one rejection in 21, so Wilson is the interval quoted). The one rejection is a fabrication finding the panel had kept unanimously - a medication-cessation reason and timeframe asserted over a transcript span the recording marks as uncertain - and the physician judged the transcript too uncertain to sustain the charge: a judgement call, counted against the instrument.

\textbf{The foils.} All nine panel-refused candidates were judged genuine failures too - two graded supporting, seven critical. This belongs with Section 4.1 rather than against the precision figure. The refusal audit above shows the panel refuses on the grounds its strict instruction sets rather than on substance, and the four-standards experiment shows a lenient reviewer verifies most of what the strict panel refuses; these foils were drawn from the 1,697 candidates the panel had itself rated high salience and refused anyway, and the panel's recorded reasons for them cite documentation convention and immateriality alike, so what the arm tests is the strict end of the standard rather than the panel's low-salience discards. The foil result is the same finding at human level - a clinician applying an ordinary reporting standard endorses both the panel's survivors and its high-importance refusals, so the sitting bounds the false-positive direction without validating the refusals as non-errors, and the published count remains the strict end of a range the physician's own standard sits well inside.

\textbf{The severity regrade.} The 20 upheld findings were regraded in the same sitting: 16 of 20 exact against the release's rubric grades, 4 adjacent, none further - and all four disagreements ran upward, the physician grading critical where the rubric said supporting. The direction is the opposite of the calibration above, where the machine graders sat about one grade more severe than the physician on the companion study's facts. The third sample, in the sitting below, runs the other way again; the three together are read at the end of this section.

The same disclosures apply: one rater, an author, one sitting, no re-test. The concealed key, the answers and the scored results are in the release; the item pack itself prints the products' notes in full and is withheld with them (Section 7).

\subsubsection{The independent clinician's sitting: a fresh sample}

Every human check above was made by one physician, who is an author of this study, the designer of the instrument they were checking and the writer of the rubric they graded against. That is the study's principal reporting shortfall, and the objection behind it is about independence, not agreement - which is what decides the design of this sitting. A second rater re-adjudicating the items the first rater already saw would add no coverage at all, and at a dozen shared items with a first-rater marginal of 29 genuine in 30 the resulting agreement coefficient is undefined, exactly zero, or negative. So the second rater was given items nobody had adjudicated.

\emph{The rater is an independent clinician: not an author of this study, and with no involvement in it at any stage.} That is what the word means everywhere it appears in this paper - independent \emph{of the study}, which is the axis the objection is about. It is not a claim about how they were found, and this paper does not describe that; nothing is claimed about arm's-length recruitment.

\textbf{Design.} Two lots of 8 items each: \textbf{6 verified findings drawn uniformly at random from the 618 and 2 foils drawn uniformly from the high-salience refused stratum}, from a fixed seed, stratified so each lot carries the same mix and shuffled within lot. Freshness is checked in code: all 45 identifiers in the precision sitting's draw are excluded and their absence asserted, leaving pools of 588 verified findings and 1,682 of the 1,697 high-salience refusals to draw from. The 16 items span 14 consultations and fall Scribe A 7, Scribe B 4, Scribe C 5; the 12 verified items span 11 consultations and fall four to each product. Each item printed the candidate's mode, description and claimed evidence quotes, then the full note and the full transcript, and asked the precision sitting's two questions in the precision sitting's words: the verdict, and on genuine verdicts only a severity grade against the same written rubric. \emph{The question wording and the rubric are copied verbatim and checked against that pack on every build}, because two raters answering differently worded questions would be measuring the wording. Each lot is one self-contained offline HTML file with no server and no network, autosaving locally, recording per-item active seconds with hidden-tab time excluded. \emph{Blinding is enforced over the emitted file, and the build fails if it is not}: no product name on a word boundary, no finding or note identifier, no status vocabulary in the file's own chrome, every item's markup skeleton byte-identical once its text is stripped, and nothing in the file that reaches the network. The read-out reported below was fixed in the manifest before the sitting ran. Both lots came back on the day they were sent.

\textbf{Precision on a fresh sample.} 12 of 12 verified findings were judged genuine, Wilson 95\% {[}75.8, 100{]}, over 11 consultations, with no rejections, no abstentions and no blanks. A consultation-clustered interval degenerates to {[}100, 100{]} at this size and stays in the released artifact, exactly as the clustered figure did in the precision sitting. \emph{This is reported beside the precision sitting's 20 of 21 and never pooled with it}: two raters, disjoint samples, and a combined ``x of y'' would be a category error. What the sitting gains is coverage and a second pair of eyes. It does not add precision: an interval running from 75.8\% to 100\% is not a precise estimate and should not be written up as one. The number of census findings a human has assessed rises from 21 to 33.

\textbf{The foils.} 4 of 4 were judged genuine, no agreement with the panel's refusals, all four graded critical. With the precision sitting's nine, thirteen high-salience refused candidates have now been put to two clinicians on disjoint samples, and every one has been judged a genuine documentation failure. All four were unanimous panel refutes and each disagreement is substantive: a meningitis teaching point written as though a non-blanching rash had been observed in a patient documented as having a blanching one; objective skin examination findings in a consultation the panel argued was not history-only; and two medication items where a clinician's speculation and an ambiguous patient ``yes'' are written up as a confirmed current medication list. \textbf{What four items can and cannot support} should be stated with the number: the four span three consultations, and the last two are the same medication exchange written up by two different products and refused twice, so the four foils hold three distinct disagreements, one of which replicates across products. With the refusal audit above and the four-standards experiment, this is the same finding at human level for the second time and with a different human: the panel refuses on the grounds its strict instruction sets rather than on substance, and it does so on a stratum it had itself rated high salience.

\textbf{The severity read, and what the three samples together now say.} Of the 12 graded pairs, \textbf{9 are exact against the rubric, 3 adjacent, none further - and all three disagreements are downward}, the clinician grading supporting where the rubric said critical. The precision sitting's four disagreements all ran upward. The two clinicians are therefore not merely differing in rate; on the same written rubric they lean in opposite directions. Two things follow, and the first is the more useful. Across both sittings, 25 of 32 rubric-versus-clinician grades are exact and no disagreement anywhere exceeds one grade, which is a real statement about the rubric: it reproduces to within one grade under two clinicians who graded disjoint material, one of them not an author. What the three samples do not establish is a direction: the disagreement is between the raters rather than between rater and rubric. \emph{No severity claim in this paper asserts a direction}, and the cluster table's severity totals should be read with that rater-dependence in mind.

\textbf{Acquiescence.} This rater answered ``genuine'' on 16 of 16 items; the physician author answered genuine on 29 of 30. Neither sitting can separate ``the flagged findings are nearly all real'' from ``clinicians asked to check flags accept them'', and on that axis this sitting is weaker than the precision sitting, not stronger, because that one contained a rejection and this one contains none. Three things bound it and none dissolves it: the foils were drawn from the refused stratum and the pack carried no status label, so the rater had no way to know which items were meant to be accepted; the severity grades are not uniform and three of them move away from the rubric, so the rater was not endorsing a template; and the items are disjoint from the precision sitting's, so this is not a second pass over the same material.

\textbf{No interassessor agreement statistic, and the reason.} Two raters on disjoint samples produce none by construction. The independent clinician was given fresh items to maximise the number of findings a human has assessed rather than to compute a coefficient that a dozen shared items could not have supported.

\textbf{Timing.} 44 minutes of active reading over 16 items: 31.4 minutes for the first lot and 12.6 for the second, median 2.0 minutes an item, fastest 24 seconds and slowest 13.9 minutes. Per-item time tracks the class of item far more than a flat rate allows: findings decidable from the note and the handed evidence quote run under 90 seconds, while the one item that genuinely required the whole transcript, an absent working diagnosis, took the longest in the sitting. A practice effect across two consecutive lots is also real here and cannot be separated from that. Any future sitting should be sized by item class rather than by a flat per-item rate.

\textbf{What this sitting does not do.} It does not discharge the separate ask for a second clinician over the \emph{judge-adjudication} sitting reported in the companion work: those items are judge outputs and these are census findings, so nothing here speaks to that comparison. It licenses no pooled figure, no agreement coefficient, and no severity direction. The concealed key, the returned answers and the scored results are in the release; the item packs themselves carry full note text and are withheld with it, as the precision sitting's pack is (Section 7).

\partdiv{Part 3 - Robustness checks}

\emph{Three ways the design itself could have produced a rate, each tested: position in the prompt, our own authored traps, and the two classes a patient record would have prefilled.}

\subsection{Position bias: no decay in discovery, early-skewed in verification}

Extraction recall is reported to fall with the depth of the target inside a long prompt \citep{chen2026graphrag}. Our discovery pass has that shape - find every instance of one failure mode inside a transcript plus a note - so if the effect operated here, every rate in the census would be an undercount weighted by position. We tested it offline from the quote spans already stored on every finding, with no additional model calls. Spans locate at \textbf{97.3\% for transcript quotes and 86.5\% for note quotes}; the rest are excluded from the position analysis and counted, never imputed.

\textbf{Discovery shows no attention-decay pattern.} Over the 13,314 candidates carrying a locatable transcript span, of the 13,678 discovery produced, mean transcript position is \textbf{0.527} against 0.5 for a uniform distribution, with the first 30\% of the transcript holding 24.5\% of those candidates and the last 30\% holding 32.9\%. That is a slight tilt towards the back of the transcript, the opposite direction to the predicted decay. The tilt is small, and it is not zero: the largest gap between the observed and the uniform cumulative distributions (the Kolmogorov-Smirnov statistic) is 0.062, which over 13,314 candidates treated as independent would not be chance, so the distribution is not flat and we do not call it so. What matters for the prediction is its direction and its length interaction, below. The test that discriminates is the length interaction, because attention decay should worsen as the transcript lengthens: mean transcript position by transcript-length quartile runs \textbf{0.519, 0.512, 0.539, 0.535} across quartile cuts at 6,573, 7,840 and 9,510 characters, with no trend across the quartiles. Whatever shape the distribution has reflects where errors occur, not how deep the model had to read.

Position within the whole prompt is early-skewed at a mean of 0.427, but the transcript occupies a fixed early portion of every prompt so that distribution is bounded by construction. It is descriptive only, its distributional test is not interpretable, and it is not evidence of anything.

\textbf{Verification skews early, and this is the real position effect.} Findings the panel upheld sit earlier than the candidate pool they came from: transcript position \textbf{0.477 for the 599 locatable verified findings against 0.527 for all candidates}, and note position \textbf{0.465 against 0.525}. With uncertainty attached - a bootstrap over the stored spans resampling whole consultations, since candidates from one consultation share a transcript - the note-position difference is -0.060 with 95\% interval {[}-0.107, -0.014{]}, and the transcript-position difference -0.050 with 95\% interval {[}-0.098, 0.000{]}, whose upper limit touches zero. So the panel's preference for early findings is supported on note position and borderline on transcript position; it is a bias in the verification layer, it points the opposite way from the discovery result, and the two are reported separately for that reason. The comparison is not adjusted for category: survival ranges from 32.2\% to 4.5\% across categories and categories are not spread evenly through a note, so part or all of the shift may be composition rather than position. The reference pool is also the full discovery output rather than the 5,898 candidates the panel reviewed, so any early skew the importance filter introduced is charged to verification here, and the note-side comparison rests on the 86.5\% of spans that locate; recomputing against the reviewed pool is the obvious next step. One product-level flag belongs with it: Scribe C's note-position distribution is the only one that skews late, at mean 0.550 with the first 30\% of the note holding 20.9\% of its findings.

\subsection{The trap-inflation check: no inflation detectable at these stratum sizes}

A design choice of ours could have produced a rate here. The authored scenarios were written to contain documentation traps, so their notes might carry more verified findings through authorship rather than scribe behaviour. The stratum that tests this was written with no knowledge of the trap schema: the authored scenarios yield 0.475 verified findings per note against the trap-blind stratum's 0.297, a difference of 0.178 whose consultation-level cluster bootstrap 95\% interval is \textbf{{[}-0.173, 0.517{]}}, over 10,000 draws and 40 consultation clusters. The interval includes zero, so no trap inflation is detected on this comparison; at these stratum sizes (120 and 37 notes) only a large effect would have shown, and with ten consultation clusters in the trap-blind stratum the interval itself is coarse rather than a precise statement of what was ruled out. The physician author separately rated two of the ten trap-blind transcripts 5 of 5 for realism, an endorsement by an interested rater rather than a measurement. By source, notes with at least one verified failure run 43.4\% on PriMock57 (99 of 228), 28.3\% on ACI-Bench (51 of 180), 18.3\% on the trap-seeded scenarios (22 of 120) and 13.5\% on the trap-blind ones (5 of 37): the public corpora produce the headline and the authored material dilutes it.

\subsection{The census net of the two classes a patient record would have prefilled}

Every product in this census was handed a transcript and nothing else: no demographics header, no problem list, no encounter date beyond what its own application supplies. Two of the clusters in A.1 are what a product carrying those fields produces when it has nothing to populate them from - \textbf{invented patient identity, 93 findings}, and \textbf{relative timing converted to invented calendar dates, 44} - and together they are \textbf{137 of the 618, 22.2\% of the census}. The Limitations state the direction of that design choice; this section states its size, so a reader can see what the census looks like with those classes removed.

\textbf{Method.} The census is recomputed in steps, each removing one more class of finding. Findings are removed and notes are not: a note keeps its place in the denominator at every step and loses only the excluded findings, so it leaves the numerator only if the excluded classes were the only thing wrong with it. Each rate carries a Wilson interval and a consultation-clustered percentile bootstrap computed by the same method and conventions as every other note-level interval in this paper - whole consultations resampled with replacement, 10,000 draws - with each row's generator seeded from the row's own tag so the results do not depend on the order the steps are computed in. Step A reuses the published tags, so its four intervals reproduce the published ones digit for digit; that is the join check, and it is asserted in code before anything new is computed.

{
{\footnotesize\renewcommand{\arraystretch}{1.18}\begin{longtable}[]{@{}
  >{\raggedright\arraybackslash}p{(\linewidth - 8\tabcolsep) * \real{0.2000}}
  >{\raggedright\arraybackslash}p{(\linewidth - 8\tabcolsep) * \real{0.2000}}
  >{\raggedright\arraybackslash}p{(\linewidth - 8\tabcolsep) * \real{0.2000}}
  >{\raggedright\arraybackslash}p{(\linewidth - 8\tabcolsep) * \real{0.2000}}
  >{\raggedright\arraybackslash}p{(\linewidth - 8\tabcolsep) * \real{0.2000}}@{}}
\toprule\noalign{}
\begin{minipage}[b]{\linewidth}\raggedright
\end{minipage} & \begin{minipage}[b]{\linewidth}\raggedright
Scribe A
\end{minipage} & \begin{minipage}[b]{\linewidth}\raggedright
Scribe B
\end{minipage} & \begin{minipage}[b]{\linewidth}\raggedright
Scribe C
\end{minipage} & \begin{minipage}[b]{\linewidth}\raggedright
\textbf{pooled}
\end{minipage} \\
\midrule\noalign{}
\endhead
\midrule\noalign{}
\endfoot
\bottomrule\noalign{}
\endlastfoot
\textbf{A. All 618 findings} (published) & 60/282 = 21.3\% {[}16.0, 27.0{]} & 57/141 = 40.4\% {[}32.6, 48.9{]} & 60/142 = 42.3\% {[}33.8, 50.0{]} & \textbf{177/565 = 31.3\% {[}27.0, 35.6{]}} \\
\textbf{B. Less invented identity} (-93) & 52/282 = 18.4\% {[}13.5, 24.1{]} & 47/141 = 33.3\% {[}25.5, 41.1{]} & 55/142 = 38.7\% {[}31.0, 46.5{]} & \textbf{154/565 = 27.3\% {[}23.1, 31.5{]}} \\
\textbf{C. Less identity and dates} (-137) & 52/282 = 18.4\% {[}13.1, 23.8{]} & 45/141 = 31.9\% {[}24.1, 39.7{]} & 43/142 = 30.3\% {[}22.5, 38.0{]} & \textbf{140/565 = 24.8\% {[}20.8, 29.0{]}} \\
\textbf{D. Bound: C plus 7 unassigned} (-144) & 52/282 = 18.4\% {[}13.5, 23.8{]} & 44/141 = 31.2\% {[}23.4, 39.0{]} & 43/142 = 30.3\% {[}22.5, 38.0{]} & \textbf{139/565 = 24.6\% {[}20.6, 28.8{]}} \\
\end{longtable}\addtocounter{table}{-1}}
}

Notes with at least one verified finding, over all 565 notes at every step; intervals resample whole consultations, and the Wilson intervals are in the released artifact. Where a step leaves a product's count unchanged, its interval can still differ in the last figure: each row is bootstrapped from its own seed, so those differences are Monte Carlo noise and not a change in the estimate.

\textbf{Read the four steps, and do not collapse them to one number.} Removing 22.2\% of the findings removes 6.5 points of the 31.3\%, a fifth of the rate, because 37 of the 177 flagged notes had nothing else wrong with them and the other 140 did. The two exclusions are deliberately kept apart, because they are not equally strong claims. A deployed scribe legitimately knows the encounter date and may resolve ``yesterday'' against it, so the date class is a weaker artefact claim than the identity class, where no product had a record to draw a name or a sex from at all. Step C is not the rate an integrated deployment would produce and must not be quoted as one: the exclusion assumes a record would have prevented every finding in both classes and nothing else. Assuming it prevents all of them removes too much, though it is close to right for identity, where a record supplies the name and sex most of those 93 findings invent; assuming it prevents nothing else removes too little, since an integrated product has more context on every axis and classes outside these two would presumably fall too. The steps are also nested subsets of one sample, so their intervals are not independent and \emph{no step may be tested against another}.

\textbf{The per-product rows are a warning about the per-product figures, and this is the more useful half of the analysis.} They are not a result about the products. The two classes fall very unevenly: identity is B 45 / C 38 / A 10 and dates are C 40 / B 4 / A 0, so the exclusion costs Scribe C 78 findings, Scribe B 49 and Scribe A 10. Scribe C falls 12.0 points and Scribe A 2.9, and the ordering inverts - at step A, Scribe C is the worst product at 42.3\% against Scribe B's 40.4\%, and at step C, Scribe B is, at 31.9\% against Scribe C's 30.3\%. Neither gap is remotely separable at these intervals, which is the point: a vendor gap measured this way is substantially a property of the instrument and of what the instrument was given, and this is a second demonstration of it that needs no change of review standard (the first is in A.3, and the census gives three further reasons for never ranking the products in Section 3.2). Anyone reading a ranking out of the per-product column has to read this table with it.

\textbf{The 55 unassigned findings are retained at every step, and two checks stop that being a hiding place.} The clustering covers 563 of the 618 and leaves 55 as noise (A.2), so they belong to no cluster and are excluded by no step. First, none of the 55 concerns an invented name or sex: a keyword sweep over all 55 descriptions raises eight hits, every one a pronoun in ordinary use, each read and cleared, and the script refuses to run if the sweep ever raises a hit that is not on the reviewed list. Second, \textbf{seven of the 55 are date-like} - a ``yesterday'' rendered as a named weekday, an invented ``couple of years ago'', and a five-finding family assigning a cessation timeframe the patient could not date - and step D removes them as an upper bound, moving the pooled rate from 24.8\% to 24.6\%. The unassigned bucket cannot change the reading. The related attribution findings in that family are deliberately not in the bound, because a patient record does not fix a misattributed cause; the seven identifiers and a reason for each are in the released artifact.

One further date-like cluster stays at every step, and the reason should be stated: the fabricated calendar year for a vague onset (cluster 9, 21 findings) turns an onset the patient could not date into a definite year, which an encounter date does not supply the way it supplies a resolution for ``yesterday'', so it is not one of the two classes a record would prefill directly. Its 21 findings rest on two consultations and four notes, none of them Scribe A's, so even treating every one of them as record-preventable could move the pooled rate by at most four notes, 0.7 points, and the per-product reading not at all beyond what step C already shows.

\partdiv{Part 4 - Reading the counts}

\emph{What is not measured, which counts look alike but answer different questions, and how many distinct errors the 618 findings represent.}

\subsection{Two categories in the scheme we did not measure}

The scheme's second tier holds thirteen categories and we hunted eleven. The two we did not are in the scheme deliberately, so that a table can show them as \textbf{not measured} rather than as zero.

\textbf{Drug and code terminology.} Grading drug names, units and codes properly needs RxNorm, SNOMED or ICD grounding that this design does not have, and the scheme's own definitions treat drug-name normalisation as correct summarising rather than as an error. The clinically consequential part of the category is covered by the dose-value pass, which we did hunt and which is the highest-surviving category in the study at 32.2\%; the one published anchor the scheme records for drug and code terminology overlaps that pass.

\textbf{Bias and stigmatising language.} Taylor's audit puts bias at 1.1\% of notes \citep{taylor2026quality}, so a dedicated pass would be expected to flag roughly six of our 565 notes. The open emergent pass remained free to surface such material, and nothing from it reached the panel under this heading. The scheme records the stigmatising-language item of the PDSQI-9 \citep{croxford2025pdsqi} - a nine-attribute instrument for grading AI-generated clinical summaries, and a separate instrument from Section 2's PDQI-9 - as the category's other published anchor.

Neither absence is a finding about the scribes. Any table putting our shares beside a published scheme's must record these two categories as not measured rather than let them read as zero; our comparator table does it in its caption, because its rows are the four published top-tier classes and these two sit a tier below them.

\subsection{Counts that look alike and answer different questions}

Three counts in this paper look like the same quantity twice: findings before and after the open pass is placed, findings against distinct errors, and cluster sizes against the top-level classes. None of the pairs should be set in adjacent columns without the sentence that separates them.

\subsubsection{Counted by the pass that found it, against counted after placement}

Every finding is labelled twice at the top tier: once by the discovery pass that found it, and once after the open emergent pass has been distributed into the published classes by the scheme's matcher. The open pass has no class of its own, so the first column lists it as a row and the second does not; in the released per-finding records the first is the \texttt{frame\_tier1} field, which is empty on open-pass findings, and the second is the count the analysis stage reports after those findings are placed.

{
{\footnotesize\renewcommand{\arraystretch}{1.18}\begin{longtable}[]{@{}
  >{\raggedright\arraybackslash}p{(\linewidth - 4\tabcolsep) * \real{0.3333}}
  >{\raggedright\arraybackslash}p{(\linewidth - 4\tabcolsep) * \real{0.3333}}
  >{\raggedright\arraybackslash}p{(\linewidth - 4\tabcolsep) * \real{0.3333}}@{}}
\toprule\noalign{}
\begin{minipage}[b]{\linewidth}\raggedright
tier 1 class
\end{minipage} & \begin{minipage}[b]{\linewidth}\raggedright
by the pass that found it
\end{minipage} & \begin{minipage}[b]{\linewidth}\raggedright
after the open pass is placed
\end{minipage} \\
\midrule\noalign{}
\endhead
\midrule\noalign{}
\endfoot
\bottomrule\noalign{}
\endlastfoot
omission & 78 & \textbf{143} \\
addition & 115 & 181 \\
wrong output & 183 & 207 \\
irrelevant or misplaced text & 46 & 46 \\
open pass, not yet placed & 196 & - \\
unmapped after placement & - & 41 \\
\textbf{total} & \textbf{618} & \textbf{618} \\
\end{longtable}\addtocounter{table}{-1}}
}

The omission row is the one that matters: \emph{78 and 143 are the same findings counted before and after the open pass is distributed}, not two estimates of how many omissions there are. Section 4's tier-1 mix, and every comparison with the published audits, uses the placed column - omission 143 of 618, 23.1\% - because that is the column whose classes match the published schemes. The pass-level column is the one to use when the question is which hunt found what, as in the survival table (Table~\ref{tab:checkability}), whose omission row counts the omission pass's own 1,284 candidates and 78 survivors.

\subsubsection{The matcher that places the open pass's findings}

The placement step above is a deterministic keyword matcher - not a model call and not a human. Eleven ordered regular-expression families (examination-not-performed, demographics, onset and timing, dropped diagnosis, safety-netting, laterality and site, fabrication, modality hardening, attribution, negation, and a generic omission catch-all) are matched against the failure-mode label the discovery model chose for the finding plus its free-text description, first match winning; each family maps to one tier-2 category and through it one tier-1 class. The matcher and its family list ship verbatim in the release, so placement is exactly reproducible. What no family's expression matches is reported as \textbf{unmapped} - 41 of the open pass's 196 survivors - and not absorbed into a neighbouring class; unmapped means the description used none of the families' vocabulary, not that the finding is unclassifiable in principle. So an open-pass finding the matcher cannot place is reported rather than absorbed into a neighbouring class. The family list is not evenhanded across the four classes, though, and it should be read alongside the counts: three families feed omission (dropped diagnosis, safety-netting and the generic catch-all), three feed addition and five feed wrong output, while none can reach irrelevant or misplaced text, and the list has no expression for a dose value or for a note contradicting itself. Of the 196 open-pass survivors it places 65 under omission, which is the difference between the two omission counts in the table above, 66 under addition, 24 under wrong output and none at all under irrelevant or misplaced text - so misplaced text is the class this placement most understates; the same matcher defines the cross-product replication cells of Section 3, there keeping its fall-through label.

\subsubsection{How many distinct errors the findings represent}

Because findings are not merged across passes, one underlying error surfaced by several hunts counts several times, and every finding-level count above inherits that. To estimate the double-counting: for each of the 132 notes carrying more than one verified finding (45 notes carry exactly one), a single model call on the panel's harsher-skeptic pin saw the full note with every finding on it and partitioned the findings into groups under one rule - findings belong together when correcting the note once would resolve all of them - instructed to keep findings separate when unsure. All 132 replies parsed. The grouping produced 220 groups over the multi-finding notes, 167 of them merging two or more findings and 53 holding one. Adding the 45 single-finding notes, \textbf{the 618 verified findings correspond to approximately 265 distinct errors}, a deduplication factor of 2.33 - 0.47 distinct errors per note against 1.09 verified findings per note. We hand-checked 24 of the 167 merged groups, the six largest and eighteen drawn at random. All 24 are defensible same-error merges, two of them borderline, none wrong. Against A.1's clusters (a group can span clusters, so these do not sum to the total): the 111 allergy and medication findings correspond to about 47 distinct errors; the 93 invented-identity findings to 49; the 53 dropped-working-diagnosis findings to 32; the 44 invented-date findings to 20; the 42 examination-provenance findings to 13; and the 11 retracted-device findings to 2. A model-free check - grouping by description-embedding similarity within each note - reads 352 to 529 distinct errors as its cosine threshold moves from 0.80 to 0.90; the model pass merges more because it recognises rephrasings across passes that embedding similarity misses. The grouping is within-note, so the same mistake appearing in two notes remains two note-level failures, matching the census's denominators. 265 is an estimate under the stated grouping rule, with the groupings released, not a count.

\subsubsection{Cluster membership against top-level counts}

The clusters are groups of similar clinical situations, not subsets of a pass. The allergy and medication cluster holds 111 findings and is placed under omission, but tier-1 omission is 143 after placement and 78 by the pass that found it, so the cluster is larger than the pass-level count of the very class it sits under. The reason is in its membership: of its 111 findings, 34 came from the open pass, 20 from the fabrication pass, 18 from the omission pass and 9 from the dose-value pass, with the remaining 30 spread across six further passes. A cluster's placement says what the group is about; it does not mean every member was found by that hunt.

Two further arithmetic facts belong with any use of the cluster table. Clusters cover \textbf{563 of the 618 verified findings}, so cluster sizes never sum to the census total. And a cluster's product columns are counts of findings, not of notes or consultations, so they cannot be turned into per-product rates; those are in Table~\ref{tab:census}.

\end{document}